\documentclass[journal]{IEEEtran}
\usepackage{amsfonts}
\usepackage{algorithmic}
\usepackage{graphicx}
\usepackage{textcomp}
\usepackage{xcolor}
\usepackage{threeparttable}
\usepackage{multirow}
\usepackage{subfigure}
\usepackage{algorithm}
\usepackage{xspace}
\usepackage{dsfont}
\usepackage{epsfig}
\usepackage{epstopdf}
\usepackage{booktabs}
\usepackage{indentfirst}
\usepackage{amsmath}
\usepackage{amsthm}
\usepackage{bm}
\usepackage{listings}
\usepackage{tabu}
\usepackage{longtable}
\usepackage{threeparttable}
\usepackage{marvosym}
\usepackage{bigstrut}
\usepackage{hyperref}
\usepackage{soul}
\usepackage{float}

\usepackage{tabularx}
\newtheorem{mydef}{Definition}

\newcommand{\name}{\home-GCL\xspace}
\newcommand{\jfe}{JFE\xspace}
\newcommand{\gf}{RFE\xspace}
\newcommand{\psa}{PSA\xspace}

\newcommand{\hgt}{HGT\xspace}
\newcommand{\home}{HOME\xspace}
\newcommand{\bj}{BJ\xspace}
\newcommand{\xa}{XA\xspace}
\newcommand{\cd}{CD\xspace}

\newcommand{\paratitle}[1]{\vspace{1.5ex}\noindent\textbf{#1}}
\newcommand{\ie}{\emph{i.e.,}\xspace}

\newcommand{\etc}{\emph{etc.}\xspace}
\newcommand{\ignore}[1]{}

\newcommand\vldbavailabilityurl{https://github.com/aptx1231/HOME-GCL}

\ifCLASSINFOpdf
\else
\fi
\begin{document}
%
\title{Jointly Learning Representations for Map Entities via Heterogeneous Graph Contrastive Learning}%
%
%


\author{Jiawei Jiang,~
        Yifan Yang,~
        Jingyuan Wang,~\IEEEmembership{Member,~IEEE},
        Junjie Wu,~\IEEEmembership{Member,~IEEE}
\IEEEcompsocitemizethanks{\IEEEcompsocthanksitem Jingyuan Wang is with the School of Computer Science and Engineering, Beihang University, Beijing 100191, China, and also with the School of Economics and Management, Beihang University, Beijing 100191, China. \protect\\
E-mail: jywang@buaa.edu.cn
\IEEEcompsocthanksitem Yifan Yang, and Jiawei Jiang are with the School of Computer Science and Engineering, Beihang University, Beijing 100191, China.
E-mail: \{yfyang, jwjiang\}@buaa.edu.cn
\IEEEcompsocthanksitem Junjie Wu is with the School of Economics and Management, Beihang University, Beijing 100191, China
}
\thanks{Corresponding Author: Jingyuan Wang.}}

%
%

\markboth{IEEE TRANSACTIONS ON INTELLIGENT TRANSPORTATION SYSTEMS}{Shell \MakeLowercase{\textit{et al.}}: Bare Demo of IEEEtran.cls for IEEE Journals}
%



\maketitle

\begin{abstract}
The electronic map plays a crucial role in geographic information systems, serving various urban managerial scenarios and daily life services. Developing effective Map Entity Representation Learning (MERL) methods is crucial to extracting embedding information from electronic maps and converting map entities into representation vectors for downstream applications. However, existing MERL methods typically focus on one specific category of map entities, such as POIs, road segments, or land parcels, which is insufficient for real-world diverse map-based applications and might lose latent structural and semantic information interacting between entities of different types. Moreover, using representations generated by separate models for different map entities can introduce inconsistencies. Motivated by this, we propose a novel method named \name for learning representations of multiple categories of map entities. Our approach utilizes a \emph{heterogeneous map entity graph} (\home graph) that integrates both road segments and land parcels into a unified framework. A \home encoder with parcel-segment joint feature encoding and heterogeneous graph transformer is then deliberately designed to convert segments and parcels into representation vectors. Moreover, we introduce two types of contrastive learning tasks, namely \emph{intra-entity} and \emph{inter-entity} tasks, to train the encoder in a self-supervised manner. Extensive experiments on three large-scale datasets covering road segment-based, land parcel-based, and trajectory-based tasks demonstrate the superiority of our approach. To the best of our knowledge, \name is the first attempt to jointly learn representations for road segments and land parcels using a unified model.
\end{abstract}


%
\IEEEpeerreviewmaketitle

\section{Introduction}

\IEEEPARstart{N}{owadays}, city Geographic Information Systems (GIS) are becoming increasingly important in citizens’ daily life and urban management by providing location-based applications, such as intelligent transportation services~\cite{libcity}, urban planning~\cite{planning}, emergency management~\cite{dgeye}, and more. \textit{Electronic Map Data}, which consist of points of interest (POIs), road segments, and land parcels, are a core component of GIS. Due to their important role in urban management and planning, it is essential to develop suitable methods to effectively characterize and model the urban electronic map data, especially in a general way.


Traditionally, electronic maps are organized using specialized special data file formats, such as Shapefile~\cite{esri1998shapefile} and GeoJSON~\cite{butler2016geojson} files, which incorporate geographic coordinate information as specialized attributes for map entities such as POIs, road segments, and land parcels. However, these file formats lack the capacity to directly model the structural relationships between map entities and thus require manual feature engineering when utilized in many downstream applications. To overcome this limitation, Map Entity Representation Learning (MERL) methods have emerged to model electronic map data~\cite{RFN, HDGE, poi2vec}. MERL methods use deep learning models to extract underlying structural and semantic information of electronic map data and convert this information into generic map entity representation vectors that can be applied in various downstream tasks, such as traffic prediction~\cite{FangQL0XZ023, guo2023self}, route planning~\cite{WangW23, XuL0X023}, and land-use classification~\cite{MVURE, MGFN}.

From a geometry perspective, map entities can be classified into three categories: points, lines, and polygons, corresponding to POIs, road segments, and land parcels, respectively. In the literature, a Map Entity Representation Learning (MERL) method is usually designed for one specific category of map entity~\cite{poi2vec, chen2020modeling, HDGE, ZE-Mob, Region2Vec, MGFN, HRNR, SARN, HyperRoad}. For POI representation learning, researchers usually leverage sequence models, such as RNNs and LSTMs~\cite{LSTM}, to model the temporal patterns from check-in sequences of POI as representations~\cite{yu2020category, kong2018hst}. Besides, skip-gram models~\cite{word2vec} are also employed to convert the POI check-in sequences for users as POI representations~\cite{poiskipgram, poi2vec}. For land parcels, early research primarily focuses on utilizing skip-gram models~\cite{word2vec, Node2Vec} or improved techniques~\cite{HDGE, ZE-Mob} to capture human mobility trajectory patterns among parcels. The subsequent studies use relation information, such as geographic distances and human mobility, to build graphs among parcels and leverage graph neural networks to generate the representation of parcels~\cite{MVURE, GEML, Region2Vec, MGFN}. For road segments, since there is a natural graph to connect segments, \ie road network, topology-aware graph embedding methods like DeepWalk~\cite{deepwalk} and Node2Vec~\cite{Node2Vec}, are widely used in the early research. Subsequently, many graph-based methods have been proposed to introduce richer information, such as topological and geospatial attribute information, into road segment representation learning~\cite{SRN2Vec, HRNR, HyperRoad}. 

\begin{figure}[t]
    \centering
    \subfigure[City Maps]{
        \includegraphics[width=0.28\columnwidth]{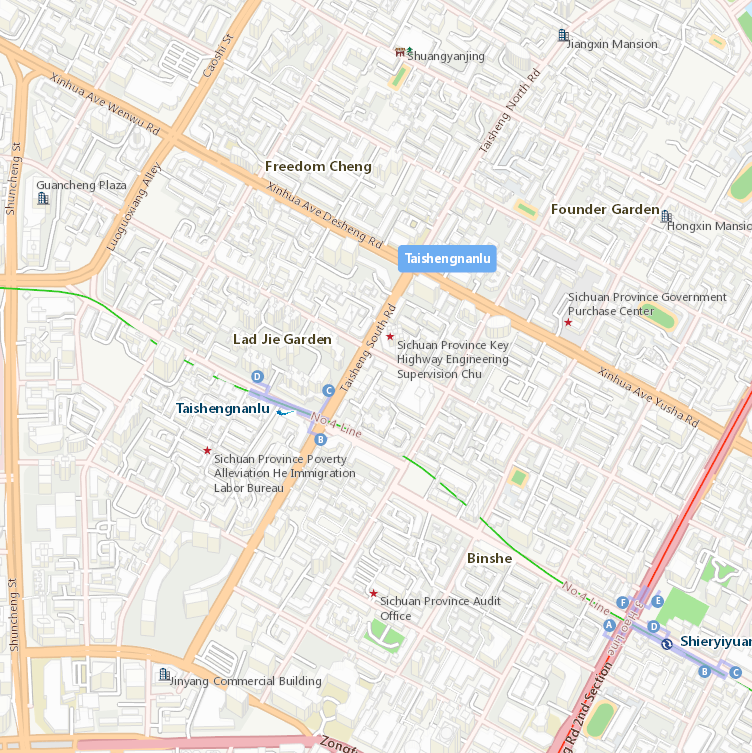}
        \label{fig:map1}
    }
    \subfigure[Road Segments]{
        \includegraphics[width=0.28\columnwidth]{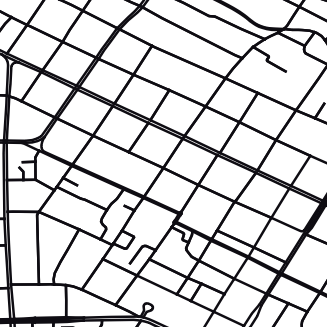}
        \label{fig:map2}
    }
    \subfigure[Land Parcels]{
        \includegraphics[width=0.28\columnwidth]{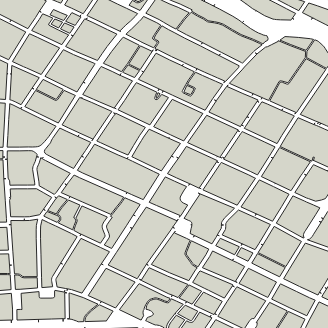}
        \label{fig:map3}
    }
    \vspace{-.1cm}
    \caption{City Maps with Corresponding Road Segments and Land Parcels in Chengdu, China.}
    \label{fig:urbanmap}
    \vspace{-.1cm}
\end{figure}

In the above-mentioned MERL studies, most of the methods are tailored for one specific category of map entities, such as POIs, road segments, or land parcels. However, as shown in Fig.~\ref{fig:urbanmap}, a real city map contains multiple categories of map entities. Only learning the representation of map entities of one type cannot satisfy the requirements of diverse and complicated real-world map-based applications. An intuitive solution is to use separate category-specific models to generate representations for map entities of different categories; this scheme, however, might cause serious consistency problems when we use the representations generated by different models but in the same application. Moreover, the interrelationships between different entity categories also contain valuable structural and semantic information. Hence, the MERL methods that target a single category of entities will inevitably lose these inter-entity relationships and thus potentially weaken the effectiveness of learned representations as well as the performance of downstream tasks. It is worth noting that some recently proposed methods suggest a joint learning scheme from a multi-level view, such as HRNR~\cite{HRNR}, which clusters road segments as structure regions and function zones and learns representations of individual segments and segment communities at the same time, but they still belong to the models for homogeneous map entity representation learning. In a nutshell, it remains a great challenge to model the complex interactive relations and structural information for heterogeneous map-entity representation learning, such as road segments and land parcels for the same map.

Driven by this motivation, we propose a novel multiple categories of map entity representation learning method, namely {\em \underline{H}eter\underline{O}geneous \underline{M}ap \underline{E}ntity \underline{G}raph \underline{C}ontrastive \underline{L}earning} (\name). In the model, we first construct a heterogeneous map entity graph (\home graph) to organize the two categories of essential map entities, \ie road segments and land parcels, where the homogeneous entity relations are modeled as three intra-entity graphs while the interactive relations between heterogeneous map entities are modeled through combine the intra-entity graphs as a heterogeneous graph. In addition, we propose a \home encoder consisting of two components, \ie a parcel-segment Joint Feature Encoding (\jfe) module and a Heterogeneous Graph Transformer (\hgt) module, to convert the segments and parcels as their representation vectors. Finally, to train the \home encoder in a self-supervised manner, we propose two types of contrastive learning tasks: intra-entity and inter-entity. The intra-entity tasks, combined with adaptive graph data augmentation, aim to learn distinguishable representations within the same category of entities (parcels or segments) by maximizing the consistency of representations between different data-augmented views. The inter-entity tasks, consisting of segment-parcel and city contrastive learning components, aim to enhance the consistency of representations for connected heterogeneous map entities, \ie segments and parcels, from a local and global perspective.

In summary, the main contributions of this paper are summarized as follows:

(1) We propose a novel map entity representation learning model named \name that can generate representation vectors of multiple categories of electronic map entities, namely road segments and land parcels, through a unified model. To the best of our knowledge, it is the first effort to use a unified model for joint learning representations of both road segments and land parcels. 

(2) We propose a new heterogeneous graph, \ie \home, to organize complex city scenarios and multi-type map entities. Regarding \home we have designed a graph encoder and two types of contrastive learning tasks specifically tailored to capture the multi-view relationships among urban entities. We believe that the \home, as well as the corresponding representation model and contrastive learning tasks, have the potential to be extended to more complex city scene modeling applications.

(3) We conduct extensive experiments to evaluate the performance of the \name framework on three large-scale datasets, encompassing three types of tasks: road segment-based, land parcel-based, and trajectory-based tasks. The results demonstrate the superiority of our approach compared to existing methods, validating the effectiveness and versatility of \name in handling multiple categories of map entity representation learning challenges.

\section{Heterogeneous Map Entity Graph}

In this section, we introduce the city map entities, then define the \textit{HeterOgeneous Map Entity Graph} (\home graph), and present the research problem addressed in this paper.

\subsection{Definitions of City Map Entities}

As shown in Fig.~\ref{fig:urbanmap}, a city map consists of a road network and land parcels divided by the road network. Given the road network of a city map, we define two categories of map entities, \ie road segments and land parcels.

\begin{mydef}[Road Segments]~\label{def:road_segment}
A road segment $s_i \in \mathcal{S}$ refers to a distinct line segment of the road network (see Fig.~\ref{fig:map2}). It is typically associated with several raw features, such as the longitude and latitude, segment type, and length, \etc The raw feature vector of the segment $s_i$ is denoted as $\bm{f}_{s_i} \in \mathbb{R}^{D_S}$, where $D_S$ is the dimension of the raw feature vectors for segments.
\end{mydef}

\begin{mydef}[Land Parcels]~\label{def:land_parcel}
A land parcel $r_i \in \mathcal{R}$ is a distinct polygon divided by the road network (see Fig.~\ref{fig:map3}). It is typically associated with raw features, such as the area size, \etc The raw feature vector of the land parcel $r_i$ is denoted as $\bm{f}_{r_i} \in \mathbb{R}^{D_R}$, where $D_R$ is the dimension of the raw feature vectors for parcels.
\end{mydef}

\subsection{Multi-view Intra-Entity Graphs}\label{nota:hdef}
In this section, we first construct three types of intra-entity graphs, namely the geographic graph, the function graph, and the mobility graph, to describe the relationships within each category of map entities.

\subsubsection{Geographic Graph} The geographic graph is used to describe the geographical relations among map entities. For road segment entities, geographical relations involve distance and topology (connections among segments). In other words, entities closer to each other should have a stronger relationship, while this relationship needs to be acted upon through connections between segments. In this way, we calculate the edge weight between segment $s_i$ and $s_j$ in the geographic graph through a piecewise function as
\begin{equation}\label{eq:road_geo}
    {e}^{S_{geo}}_{ij} =
        \left\{
        \begin{aligned}
        & \frac{1}{{\rm dist}(s_i, s_j) + \epsilon}, & \mathrm{if}~s_{i}~\mathrm{and}~ s_{j}\mathrm{~are~connected}, \\
        & 0, & {\rm otherwise},
        \end{aligned}
    \right.
\end{equation}
where ${\rm {dist}}(s_i, s_j)$ represents the Euclidean distance between the road segments $s_i$ and $s_j$. In this scenario, road segments in a city form a directed graph where segments are considered connected if they share the same intersection. To calculate the distance between two connected road segments, we utilize the latitude and longitude coordinates of the midpoint of each segment. $\epsilon$ is a small positive constant to avoid division by zero. The distance and topology information are combined by setting edge weight between unconnected segment zero. To avoid the computational problems associated with inconsistent data scales, we apply the min-max normalization to normalize the weight of connected edges between 0 and 1.

For land parcels, there are no explicit topological connections as road segments, so we directly use the distance to calculate the geographical relations. To avoid edges in the geographical graph becoming too dense, we use a threshold to filter out relationships between parcels that are too far apart. Specifically, given two land parcels $r_i$ and $r_j$, we calculate the edge weight between them as
\begin{equation}\label{eq:region_geo} 
    e^{R_{geo}}_{ij} =
        \left\{
        \begin{aligned}
        & \frac{1}{{\rm {dist}}(r_i, r_j) + \epsilon}, & {\rm {dist}}(r_i, r_j) <= \epsilon_r \\
        & 0, & {\rm otherwise},
        \end{aligned}
    \right.
\end{equation}
where ${\rm {dist}}(r_i, r_j)$ represents the Euclidean distance between the centroid of polygon $r_i$ and $r_j$, $\epsilon_r$ is the distance threshold to filter long distance edges. We also perform min-max normalization on ${e}^{R_{geo}}_{ij}$ as to the road segments.

\subsubsection{Function Graph} The function graph is used to describe functional similarity among map entities. Road segments and land parcels in different parts of a city can have different functions, such as residential areas, commercial areas, industrial areas and \etc. As reported by many studies~\cite{pois, ReMVC}, these functions can be described by points of interest (POI) around a road segment and on a land parcel. Therefore, we use the distribution of POI associated with map entities to construct the function graph edge weights.

The implementation consists of three steps. $i$) We match each POI to a corresponding map entity. For the road segment function graph, we match a POI to the road segment with the closest distance. For the land parcel function graph, we match a POI to the parcel in which it is located. $ii$) We adopt the TF-IDF model~\cite{tfidf} to construct the attribute vector for each road segment or land parcel. Specifically, we consider the category of a POI as a word, and the category set of POI is the vocabulary. For an entity, all POIs belonging to it constitute a document. We use the TF-IDF to generate a topic distribution vector for each document (entity). In this way, each entity has an attribute vector that can describe its POI semantics. $iii$) We use the cosine similarity of the attribute vectors to calculate the edge weights. We denote $\bm{p}_i$ and $\bm{p}_j$ as the POI attribute vectors of road segment $s_i$ and $s_j$ (or land parcel $r_i$ and $r_j$), the edge weight between the two entities in the function graph is calculated as
\begin{equation}\label{eq:region_sem}
    e^{S_{fun}}_{ij} = {\rm sim}\left({\bm{p}_i}, \bm{p}_j\right),
\end{equation}
where $\rm sim(\cdot, \cdot)$ is the cosine similarity function. 



\subsubsection{Mobility Graph} Human mobility in a city also reflects the relationships between map entities. We propose a mobility graph to model these relationships. In the mobility graph, edge weights are calculated using trajectory data recorded by GPS devices. A trajectory is defined as a series of latitude and longitude coordinate points. For the road segment entities, we adopt a map matching algorithm~\cite{yang2018fast} to convert all trajectories as road segment sequences, \ie segment-based trajectories~\cite{START}. Next, we calculate the transfer probabilities between road segments using segment-based trajectories. Specifically, the transfer probability (edge weights) from the segment $s_i$ to $s_j$ is calculated as
\begin{equation} \label{eq:road_mob} 
    e^{S_{mob}}_{ij} = \frac{{\rm count}\left(s_i \rightarrow s_j\right)}{\sum_k{\rm count}\left(s_i \rightarrow s_k\right)},
\end{equation}
where  ${\rm count}(s_i \rightarrow s_j)$ is the count trajectories passing $s_i$ to $s_j$. For the land parcel entities, we first build a connection for every segment to its closest parcel and then convert each segment-based trajectory to a parcel-based trajectory. Finally, we follow a similar approach to calculate the transfer probabilities ${e}^{R_{mob}}_{ij}$ for each edge of the land parcel mobility graph. For the construction of the connection between road segments and parcels, please refer to Section~\ref{nota:uhg}.

\subsubsection{Multi-view Intra-entity Graphs}
The three intra-entity graphs describe the relations with each category. Based on the intra-entity graphs, we define multi-view intra-entity graphs for road segments and land parcels.

\begin{mydef}[Multi-view Segment Graph]~\label{def:MVSG}
A multi-view segment graph is defined as $\mathcal{G}^S = (\mathcal{S}, \mathcal{E}^S)$, where $\mathcal{S}$ is a vertex set consisting of all road segments, and $\mathcal{E}^S = \{\bm{E}^{S_{geo}}, \bm{E}^{S_{fun}}, \bm{E}^{S_{mob}}\}$ is the edge set consisting of adjacency matrixes of three intra-entity graphs, \ie the edge weights $e^{R_{geo}}_{ij}$, ${e}^{S_{fun}}_{ij}$ ${e}^{S_{mob}}_{ij}$ are the $(ij)$-th item of the matrixes $\bm{E}^{S_{geo}}$, $\bm{E}^{S_{fun}}$, and $\bm{E}^{S_{mob}}$, respectively.
\end{mydef}

\begin{mydef}[Multi-view Parcel Graph]~\label{def:MVPG}
A multi-view segment graph is defined as $\mathcal{G}^R = (\mathcal{R}, \mathcal{E}^R)$, where $\mathcal{R}$ is a vertex set consisting of all land parcels, and $\mathcal{E}^R = \{\bm{E}^{R_{geo}}, \bm{E}^{R_{fun}}, \bm{E}^{R_{mob}}\}$ is the edge set.
\end{mydef}

\subsection{\home Graph and Problem Definition}\label{nota:uhg}

Besides the intra-entity graphs, we define an inter-entity graph to describe relations of map entities in different categories. In the inter-entity graph, the vertexes set is the union set of road segments and land parcels. Then, we calculate the distance between the midpoint of road segments and land parcel boundaries to assign each road segment to the nearest land parcel. If the road segment $s_i$ is assigned to the land parcel $r_j$, we set ${e}^{SR}_{ij} = 1$. Otherwise, we set it to 0. We also consider that there is a back connection from a parcel to its associated segment, \ie ${e}^{SR}_{ji} = 1$. Using ${e}^{SR}_{ij}$ as the items, we obtain an adjacent matrix $\bm{E}^{SR}$. {It is important to note that road segments are boundaries of parcels, so there is only one nearest parcel for any given road segment.}

In this way, we combine the multi-view graphs of road segments and land parcels as a \textit{HeterOgeneous Map Entity Graph} (\home graph).

\begin{mydef}[Heterogeneous Map Entity Graph]
A heterogeneous map entity graph is defined as $\mathcal{G}^H = (\mathcal{V}^H, \mathcal{E}^H, \mathcal{F}^H)$, where $\mathcal{V}^H = \mathcal{S} \cup \mathcal{R}$ is a vertex set, and $\mathcal{E}^H = \{\mathcal{E}^S, \mathcal{E}^R, \bm{E}^{SR}\}$ is the edge set, and $\bm{F}^H = \{\bm{F}^S, \bm{F}^R\}$ is the raw feature vectors for all road segments and land parcels.
\end{mydef}

The \home graph contains both node and edge heterogeneities, \ie two categories of nodes (road segments and land parcels) and seven categories of edges (three types of intra-entity edges for segment and parcels plus one type of inter-entity edge).

\paratitle{Map Entity Representational Learning.} Given the \home graph $\mathcal{G}^H = (\mathcal{V}^H, \mathcal{E}^H, \mathcal{F}^H)$,  the map entity representational learning aims to acquire an embedding function $\mathcal{F}: \mathcal{V}^H \rightarrow \mathbb{R}^{D_h}$ that project each road segment $s_i$ and land parcel $r_j$ in $\mathcal{V}^H$ to a generic, low-dimensional representation vectors $\bm{h}_{s_i} \in \mathbb{R}^{D_h}$ and $\bm{h}_{r_j} \in \mathbb{R}^{D_h}$, respectively. Here, ${D_h} \ll |\mathcal{V}^H|$ is the embedding dimension. The learned representation vectors are expected to preserve the essential characteristics within the \home graph and can be effectively employed in diverse downstream tasks, with and without fine-tuning.

\begin{figure*}[t]
    \centering
    \includegraphics[width=0.9\textwidth]{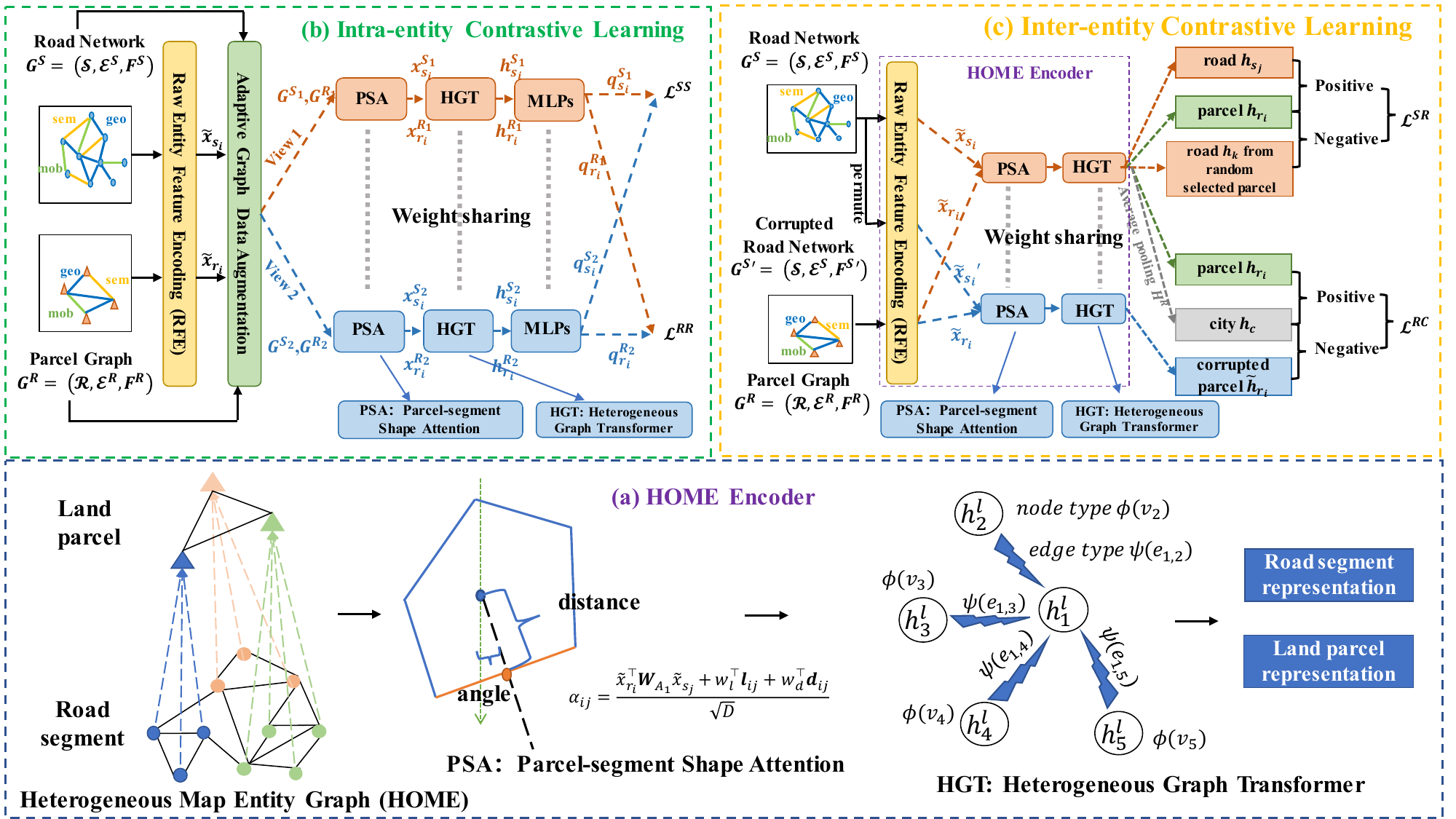}
    \vspace{-.3cm}
    \caption{Overall Framework of \name.}
    \label{fig:overall}
    \vspace{-.6cm}
\end{figure*}




\section{Representation Embedding Model}\label{model:encoder}

Fig.~\ref{fig:overall} illustrates the overall structure of the proposed \name framework in this paper, comprising three key components: (a) HOME encoder, (b) intra-entity contrastive learning, and (c) inter-entity contrastive learning. In this section, we present the model that implements the entity representation embedding function $\mathcal{F}: \mathcal{V}^H \rightarrow \mathbb{R}^{D_h}$, referred to as the \home encoder, represented in part (a) of the figure. The \home encoder comprises two main modules: the Parcel-Segment Joint Feature Encoding (\jfe) module and the Heterogeneous Graph Transformer (\hgt) module.



\subsection{Parcel-segment Joint Feature Encoding} \label{model:fea}

\subsubsection{Raw Entity Feature Encoding}  \label{sec:feature_encoding}
As defined in Definition~\ref{def:road_segment} and~\ref{def:land_parcel}, each road segment/land parcel has a raw feature vector, \ie $\bm{f}^S \in \mathbb{R}^{D_S}$ and $\bm{f}^R \in \mathbb{R}^{D_R}$, respectively. The \jfe module first encodes these raw features as dense embedding vectors through a linear embedding layer.


Then, for the segment $s_i$ and the parcel $r_i$, we denote their raw feature embedding vectors as $\{\bm{u}_1, \ldots, \bm{u}_{D_S}\}$ and $\{\bm{v}_1, \ldots, \bm{v}_{D_R}\}$. Then, we concatenate the raw feature embedding vectors of an entity as a long vector and adopt a multilayer perceptron network to compress the long vectors as a dense entity feature embedding vector as
\begin{equation}\label{eq:embed_vec}
    \tilde{\bm{x}}_{s_i} = \mathrm{MLP}\left(\bm{u}_1\Vert\ldots\Vert\bm{u}_{D_S} \right), \tilde{\bm{x}}_{r_i} = \mathrm{MLP}\left(\bm{v}_1\Vert\ldots\Vert\bm{v}_{D_R}\right),
\end{equation}
where $\mathrm{MLP}(\cdot)$ is the multilayer perceptron network, which contains two fully connected layers. $\Vert$ denotes the concatenation operation. $\tilde{\bm{x}}_{s_i} \in \mathbb{R}^D$ and $\tilde{\bm{x}}_{r_i} \in \mathbb{R}^D$ are the dense embedding vectors for the segment $s_i$ and the parcel $r_i$. We set embedding vectors of all map entities with the uniform dimension $D$.

\subsubsection{Parcel-segment Shape Attention}\label{sec:shape} Besides the raw features, the geometric structure is also an important feature of the map entities. These structures involve both road segments and land parcels, \ie the road segments surround the boundaries of land parcels. To model this special feature, we propose the shape attention to enhance feature embedding vectors generated by Eq.~\eqref{eq:embed_vec}.

Specifically, given a land parcel $r_i$ and a road segment $s_{j}$ that is assigned to $r_i$, \ie in the \home graph ${e}^{SR}_{ij} = 1$, we calculate the attention score between the land parcel and the road segment as
\begin{equation}\label{eq:att}
    \mathrm{ATT}_{ij} = \frac{\exp(\alpha_{ij})}{\sum_{s_{j'} \in \mathcal{S}_{r_i}}\exp(\alpha_{ij'})},\;\;\mathrm{where}\;\;\;
    \alpha_{ij} = \frac{{\tilde{\bm{x}}_{r_i}^\top \bm{W}_{A_1} \tilde{\bm{x}}_{s_{j}}}}{\sqrt{D}},
\end{equation}
where $ \mathcal{S}_{r_i}$ is the set of segments that are assigned with $r_i$ in the \home graph, and $\bm{W}_{A_1} \in \mathbb{R}^{D\times D}$ is a learnable parameter. The attention score function~\eqref{eq:att} only considers the connection relationship between segments and parcels. To model the geometric relations between a parcel and the surrounding segments, we introduce two shape features into the attention scores: $i$) the distance from the centroid of the parcel to the segments; $ii$) the direction angle of the line from the road segment's center to the parcel's centroid. In this way, the $\alpha_{ij}$ in Eq.~\eqref{eq:att} is improved as
\begin{equation}\label{eq:att_enchance}
    \alpha_{ij} = \frac{{\tilde{\bm{x}}_{r_i}^\top \bm{W}_{A_1} \tilde{\bm{x}}_{s_{j}} + \bm{w}_l^\top \bm{l}_{ij} + \bm{w}_d^\top \bm{d}_{ij}}}{\sqrt{D}}, 
\end{equation}
where $\bm{l}_{ij}\in \mathbb{R}^{D_l}$ and $\bm{d}_{ij}\in \mathbb{R}^{D_d}$ are embedding vectors for the distance between centroid of $r_i$ to $s_j$ and the direction angle of $r_i$ and $s_j$, and $\bm{w}_l\in \mathbb{R}^{D_l}$ and $\bm{w}_d\in \mathbb{R}^{D_d}$ are learnable parameters. By incorporating the two terms, our model effectively captures the shape feature of the parcel polygons and their relative positional relationship with the surrounding road segments. 

Using $\mathrm{ATT}_{ij}$ with the $\alpha_{ij}$ calculated by Eq.~\eqref{eq:att_enchance} as weights, we combine feature embedding vectors of all segments that assigned to $r_i$ as
\begin{equation}\label{eq:combine}
     {\bm{a}}_{r_i} = \sum_{s_{j} \in \mathcal{S}_{r_i}} \mathrm{ATT}_{ij} \bm{W}_{A_2} \tilde{\bm{x}}_{s_{j}},
\end{equation}
where $\bm{W}_{A_2} \in \mathbb{R}^{D\times D}$ is a learnable parameter matrix. Finally, we use ${\bm{a}}_{r_i}$ to enhance the feature embedding vectors of parcels $\tilde{\bm{x}}_{r_i}$ as
\begin{equation}\label{eq:ssa}
    \bm{x}_{r_i} =  \bm{W}_{A_3} \bm{a}_{r_i} + \tilde{\bm{x}}_{r_i},
\end{equation}
where $\bm{W}_{A_3} \in \mathbb{R}^{D \times D}$ represents the learnable parameters. Since the segments do not have the geometric shape as land parcels, we directly set $\bm{x}_{s_i} = \tilde{\bm{x}}_{s_i}$ in subsequent modules of our model.

\subsection{Heterogeneous Graph Transformer}\label{sec:hgt}

Given a \home graph $\mathcal{G}^H$ with its node's feature embedding vectors, we use a Heterogeneous Graph Transformer (\hgt)~\cite{HGT} to generate comprehensive representations for map entities. The \hgt module adopts a node encoder $f_n(\cdot)$ and an edge encoder $f_e()$ to handle the heterogeneity of nodes and edges. Given a node $v_i$, which can be a road segment or a land parcel, we indicate its category as $\phi(v_i)$, and given an edge $e_{ij}$, we indicate its types as $\psi(e_{ij})$. 

In the \home graph, the edges have seven categories, including segment-geographic-segment, segment-function-segment, segment-mobility-segment, parcel-geographic-parcel, parcel-function-parcel, parcel-mobility-parcel, and parcel-near-segment. The \hgt module implements the node encoder and the edge encoder as
\begin{equation}\label{eq:hgt_encoder}
    \begin{aligned}
        f_n(v_i) & =  \bm{W}_{\phi(v_i)} \bm{x}_{v_i}, \\
        f_e(e_{ij}) & =  e_{ij}\cdot \bm{W}_{\psi(e_{ij})} f_n(v_j),
    \end{aligned}
\end{equation}
where $\bm{W}_{\phi(v_i)}$, $\bm{W}_{\psi(e_{ij})} \in \mathbb{R}^{D \times D}$ are learnable parameters, which are distinct for different categories of segments and parcels.

To model the impacts of neighboring heterogeneous nodes from heterogeneous edges to a target node, the \hgt module employs a dot-product attention mechanism. The attention first encodes the target node $v_i$ as a query vector and encodes the edges $e_{ij}$ connected to $v_i$ as key and value vectors, \ie
\begin{equation}\label{eq:hgt_eqkv}
    \bm{q}_{v_i} = \bm{W}_q f_n(v_i), \; \; \bm{k}_{e_{ij}} = \bm{W}_k f_e(e_{ij}),  \; \;\bm{v}_{e_{ij}} = \bm{W}_v f_e(e_{ij}),
\end{equation}
where $\bm{W}_q, \bm{W}_k, \bm{W}_v \in \mathbb{R}^{D \times D}$ are learnable parameter matrices. Then, the attention score for the neighbor node $v_j$ is calculated as
\begin{equation}\label{eq:hgt_score}
    \alpha_{ij} = \frac{\exp\left(\bm{q}_{v_i}^\top\bm{k}_{e_{ij}}/{\sqrt{D}}\right)}{\sum_{v_{j'} \in \mathcal{N}_i}\exp\left(\bm{q}_{v_i}^\top\bm{k}_{e_{ij'}}/{\sqrt{D}}\right)},
\end{equation}
where $\mathcal{N}_t$ is the neighboring nodes set of $v_i$. Then, the output representation of target node $v_i$ is obtained by aggregating the value vectors of neighboring edges as
\begin{equation}\label{eq:hgt_out}
    \bm{h}_{v_i} ={\sum_{v_{j} \in \mathcal{N}_i}{\alpha_{ij} \bm{v}_{e_{ij}}}}+ f_n(v_i),
\end{equation}
where $\bm{h}_{v_i}$ is the output representation vector for the target node $v_i$. We also use the multi-head attention to stabilize the training process, where the final output is the average of all attention heads.

\section{Contrastive Learning Strategy}

In this section, we introduce two self-supervised contrastive learning tasks, namely inter-entity contrastive learning and intra-entity contrastive learning. These tasks are employed to train the \home encoder, depicted in part (b) and (c) of Fig.~\ref{fig:overall}.

\subsection{Intra-entity Contrastive Learning}
The function of intra-entity contrastive learning is to inject underlying patterns of entities within the same category (parcel or segment) into the entity representations.

\subsubsection{Adaptive Graph Data Augmentation} In the intra-entity contrastive learning, we use two data augmentation methods, namely edge augmentation and node feature augmentation, to generate a data-augmented version for the multi-view intra-entity graphs (see Definition~\ref{def:MVSG} and~\ref{def:MVPG}). These data-augmented graphs have similar underlying patterns to the original intra-entity graphs but are injected with some randomness in the graph structure and features.

\paratitle{Edge Augmentation.} The edge augmentation randomly removes a portion of edges from the multi-view intra-entity graphs to generate data-augmented graphs. To preserve the underlying patterns, we use weight to indicate the edge's importance as well as set a low-importance edge with a high removing probability, and vice versa. Using geographic edges of segments as an example, the removing probability of the edge from $s_i$ to $s_j$ is set as
\begin{equation}\label{eq:pedge}
    p^{S_{geo}}_{ij} = \min\left\{\left(1 - e^{S_{geo}}_{ij}\right) \cdot p_e, \ p_{\tau}\right\},
\end{equation}
where $e^{S_{geo}}_{ij}$ represents the edge weight calculated according Eq.~\eqref{eq:road_geo}, which ranges between 0 and 1. The hyperparameter $p_e$ controls the overall probability of edge removal, while $p_{\tau} = 0.7$ truncates probabilities to preserve the graph structure by avoiding excessively high removal probabilities. In this way, the data-augmented graphs generated by Eq.~\eqref{eq:pedge} maintain the important structure of the original segment geographic graph while involving some random variation on the edges. For other categories of inter-entity edges, we adopt the same removing methods to augment data.

\paratitle{Node Feature Augmentation.} The node feature augmentation approach randomly masks items in the node's representation vectors to generate data-augmented graphs. Similar to edge augmentation, we set different mask probabilities according to the importance of a feature. We denote $x_{s_i}^{(k)}$ as the $k$-th feature of the segment $s_i$'s representation vector. Its importance is measured by the weighted average of the $k$-th feature for all segments. The average weight is set as the in-degree of each node. Specifically, the importance of the $k$-th feature, denoted as ${c}_k$, is set as
\begin{equation}\label{eq:wnode}
    {c}_k = \sum_{s_i \in \mathcal{S}} |{x}_{s_i}^{(k)}| \cdot \mathrm{Degree}\left({s_i}\right),
\end{equation}
where $x_{s_i}^{(k)}$ represents the $k$-th dimension of the representation vector $\bm{x}_{s_i}$, and $|\bm{x}^S_{i,j}|$ denotes the absolute value of that dimension to capture the magnitude of the feature value. $\mathrm{Degree}\left({s_i}\right)$ is a function to return the in-degree of the segment $s_i$ in the multi-view intra-entity graphs, including three different types of edges.
 
We introduce in-degree as a weight due to the fact that higher in-degree road segments are usually considered transportation hubs and have higher importance. To ensure that the weight $c_k$ is between 0 and 1, we apply a min-max normalization. Then, we determine the probability of the $k$-th feature being masked as
\begin{equation}\label{eq:pnode}
    {p}_{k} = \min\Big\{\left(1 - \bm{c}_k\right) \cdot p_n, \ p_{\tau}\Big\},
\end{equation}
where $p_n$ controls the overall probability of feature masking. This equation ensures that the higher the weight, the lower the probability of a feature being masked. For the features of land parcels, the same mask approach is applied to generate the augmented data.

\subsubsection{Intra-entity Contrastive Learning Loss}

The loss function of intra-entity contrastive learning encourages the model to learn underlying representations of map entities by maximizing the consistency of representations between different data-augmented graphs. We also use the road segment as an example to explain the implementation. Given a multi-view segment graph $\mathcal{G}^S$,  we perform two rounds of data augmentation to obtain two different data-augmented graphs ${\mathcal{G}}^{S_1}$ and ${\mathcal{G}}^{S_2}$. Next, we input the two data-augmented views into a parameter-shared \home encoder defined in section~\ref{model:encoder} and generate corresponding segment representations, denoted as $\bm{h}^{S_1}_{s_i}$ and $\bm{h}^{S_2}_{s_i}$ for the segment $s_i$. Then, we map these representations into a low-dimensional space using a projection head as
\begin{equation}\label{eq:phead}
    \bm{q}^{S_1}_{s_i} = \bm{w}_2^\top\left({\rm ELU}\left(\bm{w}_1^\top \bm{h}^{S_1}_{s_i}\right)\right), \;\;\; \bm{q}^{S_2}_{s_i} = \bm{w}_2^\top\left({\rm ELU}\left(\bm{w}_1^\top \bm{h}^{S_2}_{s_i}\right)\right),
\end{equation}
which is a two-layer feed-forward neural network (FNN) with an ELU activation~\cite{elu} in the middle. The parameters of the projection head are shared between both augmented versions.

In loss function, we consider the pair of $\bm{q}^{S_1}_{s_i}$ and $\bm{q}^{S_2}_{s_i}$ for the same segment as positive samples, and pair of different segments as negative samples. Then, the loss function for the positive sample pair of $s_i$ is defined as
\begin{equation}\label{eq:road-road-loss} \small
\begin{split} 
    & \mathcal{L}^{SS}\left(\bm{q}^{S_1}_{s_i}, \bm{q}^{S_2}_{s_i}\right)= \\
    & -\log \frac{\exp\left({\rm sim}(\bm{q}^{S_1}_{s_i}, \bm{q}^{S_2}_{s_i})/\tau\right)}{ \displaystyle\sum_{s_j \in \mathcal{S}}\exp\left({\rm sim}(\bm{q}^{S_1}_{s_i}, \bm{q}^{S_2}_{s_j})/\tau\right) + \displaystyle\sum_{s_j \neq s_i }\exp\left({\rm sim}(\bm{q}^{S_1}_{s_j}, \bm{q}^{S_1}_{s_j})/\tau\right)},
\end{split}
\end{equation}
where $\tau$ is a temperature hyperparameter and ${\rm sim(\cdot, \cdot)}$ is the cosine similarity function. The final segment intra-entity contrastive loss $\mathcal{L}^{SS}$ is obtained by averaging the losses for all positive pairs. The intra-entity contrastive loss for parcels is computed in the same way and denoted as $\mathcal{L}^{RR}$.

\subsection{Inter-entity Contrastive Learning}

The function of inter-entity contrastive learning is to learn consistent representations for the entities that have close relationships but are within different categories. This mechanism allows the representation of different categories of map entities to be brought under the same system. The inter-entity contrastive encompasses two aspects: the segment-parcel contrastive and the city contrastive, which perform contrastive learning from local and global perspectives, respectively.



\subsubsection{Segment-parcel Contrastive Learning} The segment-parcel contrastive learning aims to learn consistent representations between a connected segment-parcel pair in the \home graph. This aim is achieved by maximizing the mutual information between parcel representations and road segment representations connected to the land parcel. For the parcel $r_i$, we denote $\mathcal{S}_{r_i}$ as a positive sample set consisting of segments connected with $r_i$, while $\mathcal{S}_{r_i'}$ consisting of segments that are connected with another randomly selected parcel $r_i'$ is considered as a negative sample set. Then, the segment-parcel contrastive loss is defined as
\begin{equation}\label{eq:local_loss}
    \begin{split}
    \mathcal{L}^{SR} = &-\left(\frac{1}{N_{pos}}\sum_{r_i \in \mathcal{R}}\sum_{s_j \in \mathcal{S}_{r_i}}\log \mathcal{D}\left(\bm{h}_{r_i}, \bm{h}_{s_j}\right) \right. \\
    &\;\;\;\;\left. + \frac{1}{N_{neg}}\sum_{r_i \in \mathcal{R}}\sum_{s_k \in \mathcal{S}_{r_i'}}\log\Big(1 - \mathcal{D}\left(\bm{h}_{r_i}, \bm{h}_{s_k}\right)\Big)\right),
    \end{split}
\end{equation}
where $N_{pos} = \sum_{r_i\in \mathcal{R}}|\mathcal{S}_{r_i}|$ and $N_{neg} = \sum_{r_i\in \mathcal{R}}|\mathcal{S}_{r_i'}|$ are number of the positive samples and the negative samples, respectively. $\mathcal{D}(\bm{h}_{s_i}, \bm{h}_{r_j}) = \mathrm{Sigmoid}(\bm{h}_{s_i}^\top \bm{W}_D\bm{h}_{r_j})$ is function to calculates similarity between the representations of a segment and parcel pair, where $\bm{W}_D$ is a learnable parameter matrix. We did not directly use the cosine similarity here because representations of segments and parcels can not be perfectly consistent. A learnable parameter matrix can offer more flexibility for the similarity measure. The segment-parcel contrastive learning enables the representations to distinguish similar and dissimilar parcel-segment pairs, which enhances the model's ability to capture the interactive relationship of map entities in different categories.

\subsubsection{City Contrastive Learning} The city contrastive learning provides a mechanism for our model to learn a global relationship among map entities. To achieve this, we introduce a visual global city node to the \home graph. We assume the city node connects all parcel nodes of the \home graph. The representation of the city node is calculated as a simple average pooling of the parcel representations, \ie $\bm{h}_{c} = \frac{1}{|\mathcal{R}|} \sum_{r_i \in \mathcal{R}} \bm{h}_{r_i}$.

The task of city contrastive learning is to maximize the mutual information between the city node representation and parcel representations. We consider the city representation and a parcel representation as a pair of positive samples, \ie $<\bm{h}_{c}, \bm{h}_{r_i}>$. Besides, we also construct a set of fake representations for the parcels as negative samples. Specifically, we corrupt the raw feature matrix $\bm{f}_{s_i}$ for all road segments by randomly permuting its rows, resulting in a set of fake road segments. Then, we feed the fake road segment feature into the parcel-segment shape attention to obtain fake representations for all parcels, denoted as $\tilde{\bm{h}}_{r_i}$. The representation pairs $<\bm{h}_{c}, \tilde{\bm{h}}_{r_i}>$ are used as negative samples. The loss function for city contrastive learning is defined as
\begin{equation}\label{eq:global_loss} 
    \begin{aligned}
    \mathcal{L}^{C} = -\left(\frac{1}{|\mathcal{R}|}\sum_{r_i \in \mathcal{R}}\log \mathcal{D}\left(\bm{h}_{r_i}, \bm{h}_{c}\right) \right. \\ 
    \left. + \frac{1}{|\mathcal{R}|}\sum_{r_i \in \mathcal{R}}\log\left(1 - \mathcal{D}\left(\tilde{\bm{h}}_{r_i}, \bm{h}_{c}\right)\right)\right),
    \end{aligned}
\end{equation}
where $\mathcal{D}(\bm{h}_{r_i}, \bm{h}_{c}) = \mathrm{Sigmoid}(\bm{h}_{r_i}^\top \bm{W}_C\bm{h}_{c})$ and $\bm{W}_C$ is a learnable parameter matrix. In the loss function Eq.~\eqref{eq:global_loss}, the model learns to distinguish between similar and dissimilar parcel-city pairs. This setup lets representations learned by our model also contain global relationships among map entities.

In contrastive learning, we use perturbing the representations of segments to generate negative parcel samples instead of directly perturbing the representations of the parcel. The reason has two aspects: First, the city representation $\bm{h}_c$ obtained through the average pooling of $\bm{f}_{r_i}$, so directly perturbing $\bm{f}_{r_i}$ as negative samples make the task of distinguishing real/fake parcels too easy. Second, this setup involves segment representations segments into the city contrastive learning in an indirect way, which makes our method learn more comprehensive global relationships among map entities.

\subsection{Loss Fusion}
Finally, we pre-train the \home encoder with both intra-entity and inter-entity contrastive tasks. The training loss is defined as
\begin{equation}
    {\mathcal{L}} = \lambda_1 {\mathcal{L}}^{SS} + \lambda_2 {\mathcal{L}}^{RR} + \lambda_3 {\mathcal{L}}^{SR} + \lambda_4 {\mathcal{L}}^{C},
\end{equation}
where $\lambda_*$ are the hyperparameters to balance these tasks. By combining inter-entity and intra-entity contrastive learning, our model can comprehensively understand the urban heterogeneous map entity graph and its intricate relationships. 




\makeatletter
\newcommand{\thickhline}{%
    \noalign {\ifnum 0=`}\fi \hrule height 1pt
    \futurelet \reserved@a \@xhline
}

\section{Experiments}

In this section, we conduct comprehensive experiments to evaluate the performance of the \name on three large-scale datasets.

\subsection{Datasets and Preprocessing}\label{exp:dataset}
We utilize three real-world and large-scale datasets from three cities: Beijing (\bj), Xi'an (\xa), and Chengdu (\cd). Each dataset comprises four types of data, namely point of interest data, road network data, land parcel data, and trajectory data. The details of each data type are as follows:

(1) The \textbf{Point of Interest (POI) Data} is obtained from Baidu Maps~\footnote{\url{https://lbsyun.baidu.com/}}. POI data includes more than 10w data collected in 3 cities in 20 major categories and 244 minor categories, enabling the computation of functional information among urban entities. 


(2) The \textbf{road network data} is obtained from OpenStreetMap~\footnote{\url{https://openstreetmap.org/}}~\cite{OpenStreetMap}. We collect various features for each segment, including its category, length, number of lanes, maximum speed, longitude, and latitude.


(3) The \textbf{land parcel data} is sourced from Beijing City Lab~\footnote{\url{https://www.beijingcitylab.com/data-released-1/}}~\cite{bj_dataset}. This dataset divides the city into multiple land parcels, where parcels are the basic spatial units divided by the road network. The considered features include the dominant function of the parcel, whether the parcel is in the central business district (CBD), the number of buildings, the average floor number, the area of the parcel, as well as its longitude and latitude.


(4) The \textbf{trajectory data} consists of GPS trajectory data collected from taxis in the three cities. The trajectory data for Beijing was collected in November 2015 by taxis, released by~\cite{START}. Moreover, the trajectory data for Xi'an and Chengdu was collected in November 2018 and obtained from the DiDi GAIA~\footnote{\url{https://outreach.didichuxing.com/research/opendata/en}} project.

\textbf{Data preprocessing}: We preprocess the GPS trajectories by performing map matching~\cite{yang2018fast} to align them with the road segments, resulting in road segment-based trajectories. For the \xa and \cd datasets, we split the trajectories chronologically with a ratio of 6:2:2 to create the training, validation, and test subsets. The \bj trajectories are split into training, validation, and test subsets in chronological order, covering 18, 5, and 7 days, respectively, with fewer data available on November 25~\cite{START}. The same data partitioning method is used for both pre-training and fine-tuning phases. The statistics for the three datasets are presented in Tab.~\ref{tab:data_detail}, where the symbol `$\#$' denotes the number of instances for each corresponding dataset feature. The codes and datasets can be found here~\footnote{\url{\vldbavailabilityurl}}. 

\begin{table}[t]
  \centering
  \caption{Statistics of the Three Datasets.}
  \vspace{-0.1cm}
    \begin{tabular}{c|c|c|c}
    \thickhline
    Dataset & \bj    & \xa    & \cd \\
    \hline
    Time span & 2015.11 & 2018.11 & 2018.11 \\
    \hline
    \# Trajectory & 1018312 & 384618 & 559729 \\
    \hline
    \# Usr & 1677  & 26787 & 48295 \\
    \hline
    \# POI & 81180 & 19107 & 17300 \\
    \hline
    \# road segments & 40306 & 5269  & 6195 \\
    \hline
    \# land parcels & 11208 & 1056  & 1306 \\
    \hline
    max\_len & 128   & 64    & 64 \\
    \thickhline
    \end{tabular}%
  \label{tab:data_detail}%
  \vspace{-0.3cm}
\end{table}%

\subsection{Downstream Tasks}\label{exp:tasks}
We conduct experiments on the following three types of downstream tasks to evaluate the learned representations:

\subsubsection{Road Segment-based Tasks} Following previous work~\cite{HRNR, Toast, JCLRNT}, we design the following three tasks:

(1) Road classification task: This task assigns road types as labels, considering only the five most frequent types. We employ \textit{a fully connected layer with softmax activation} as a multi-classifier based on the representation $\bm{h}_{s_i}$.

(2) Traffic flow prediction task: This task aims to predict the inflow and outflow of road segments. We use \textit{a ridge regression model}~\cite{ridge} to predict road segment inflow and outflow based on the representation $\bm{h}_{s_i}$.

(3) OD flow prediction task: This task focuses on predicting the origin-destination (OD) flow of road segments. We use \textit{a bilinear model}~\cite{GEML} to estimate the OD flow between $s_i$ and $s_j$ using the representations $\bm{h}_{s_i}$ and $\bm{h}_{s_j}$.


\subsubsection{Land Parcel-based Tasks}
Similar to the road segment-based tasks and following previous works~\cite{MVURE, MGFN}, we have designed three downstream tasks to thoroughly evaluate our parcel representations: parcel classification, traffic flow prediction, and OD flow prediction. The experimental setups for these tasks are also similar to the road segment-based tasks.

\subsubsection{Trajectory-based Tasks}
In addition, we also explore the effectiveness of the model-learned road segment representations on trajectory-based tasks, following previous work~\cite{SARN, JCLRNT, Toast}. We introduce two trajectory-based tasks, for which we apply map matching~\cite{yang2018fast} to convert GPS trajectories into road segment-based trajectories, enabling these tasks.

To compare performance, we utilize two types of models. The first class is the road network representation model, where we use \textit{a two-layer LSTM~\cite{LSTM}} to process the learned road segment representations and obtain the trajectory representation $\bm{t}_i$ by averaging the hidden states of the LSTM. Subsequently, we feed $\bm{t}_i$ into task-specific downstream components to obtain predicted values for the trajectory-based tasks. The second class is the self-supervised trajectory representation models, which are directly pre-trained to obtain the trajectory representation $\bm{t}_i$, eliminating the need for the LSTM. Thus, $\bm{t}_i$ can be directly input into the downstream components, and both upstream and downstream models can be fine-tuned simultaneously to achieve better performance. The two trajectory-based tasks are as follows:

(1) Travel time estimation task: This task estimates the travel time from the origin to the destination based on a given trajectory. We employ \textit{a fully connected layer} to predict the travel time using the representation $\bm{t}_i$.

(2) Trajectory similarity search task: This task measures the similarity between two trajectories. We calculate the L1 distance between two trajectory embeddings as the predicted similarity. We use the Fréchet distance~\cite{alt1995computing} as the ground truth similarity metric between two trajectories. The model minimizes the mean squared error (MSE) between the predicted similarity and the ground truth. 

\subsection{Baselines}
We compare our approach with 21 representation learning methods as baselines, falling into four categories: (1) Graph embedding methods, including DeepWalk~\cite{deepwalk}, Node2Vec~\cite{Node2Vec}, LINE~\cite{LINE}, and DGI~\cite{DGI}; (2) Road network representation learning methods, including RFN~\cite{RFN}, SRN2Vec~\cite{SRN2Vec}, HRNR~\cite{HRNR}, Toast~\cite{Toast}, JCLRNT~\cite{JCLRNT}, HyperRoad~\cite{HyperRoad}, and SARN~\cite{SARN}; (3) Land parcel representation learning methods, which consist of HDGE~\cite{HDGE}, ZE-Mob~\cite{ZE-Mob}, MV-PN~\cite{MV-PN}, MVURE~\cite{MVURE}, Region2Vec~\cite{Region2Vec}, MGFN~\cite{MGFN}, and ReMVC~\cite{ReMVC}; and (4) Trajectory representation learning methods, where we utilize baselines used for road network representations and combine them with LSTM models for downstream tasks. Additionally, we also include trajectory representation learning models for comparison, including t2vec~\cite{t2vec}, trajectory2vec~\cite{trajectory2vec}, and Trembr~\cite{Trembr}.

\subsection{Experimental Settings}\label{exp:set}

\subsubsection{Parameter Settings}
All experiments are conducted on Ubuntu 20.04 using an NVIDIA GeForce 3090 GPU. We implement our \name and all baselines using PyTorch 2.0.1~\cite{pytorch}. The embedding size $D$ for all methods is set to 128. The number of layers in the \hgt module is 2, and the number of attention heads $H$ is 8. The dropout ratio is 0.2, and the temperature parameter $\tau$ is 0.4. For data augmentation, we control the truncates probability $p_{\tau}$ to be 0.7, the edge deletion probability $p_e$ to be 0.3, and the feature mask probability $p_n$ to be 0.4 when generating view 1. For generating view 2, we set the three probabilities to 0.7, 0.4, and 0.3. We consider both inter-entity contrastive and intra-entity contrastive losses equally important and, therefore, set all $\lambda_*$ to 0.25, giving each of the four tasks the same weight. We use the Adam optimizer~\cite{Adam} with a learning rate of 0.01 and train the models for 1000 epochs. For baselines that require input features for road segments and land parcels, we process the features in the same way as our \name to maintain a fair comparison. 

\begin{table*}[t]
  \centering
  \caption{Performance on Road Segment-based Tasks.}
  \vspace{-0.2cm}
 \resizebox{\textwidth}{!}{
    \begin{tabular}{c|cc|cc|cc|cc|cc|cc|cc|cc|cc}
    \thickhline
    Data  & \multicolumn{6}{c|}{BJ}                       & \multicolumn{6}{c|}{XA}                       & \multicolumn{6}{c}{CD} \bigstrut\\
    \hline
    Tasks & \multicolumn{2}{c|}{Classification} & \multicolumn{2}{c|}{Flow prediction} & \multicolumn{2}{c|}{OD prediction} & \multicolumn{2}{c|}{Classification} & \multicolumn{2}{c|}{Flow prediction} & \multicolumn{2}{c|}{OD prediction} & \multicolumn{2}{c|}{Classification} & \multicolumn{2}{c|}{Flow prediction} & \multicolumn{2}{c}{OD prediction} \bigstrut\\
    \hline
    models\textbackslash{}metrics & Mi-F1 & Ma-F1 & MAE   & RMSE  & MAE   & RMSE  & Mi-F1 & Ma-F1 & MAE   & RMSE  & MAE   & RMSE  & Mi-F1 & Ma-F1 & MAE   & RMSE  & MAE   & RMSE \bigstrut\\
    \hline
    DeepWalk & 0.364  & 0.146  & 1.333  & 1.942  & 0.227  & 0.407  & 0.441  & 0.318  & 2.103  & 3.246  & 2.642  & 7.810  & 0.339  & 0.278  & 2.168  & 4.156  & 2.809  & 15.106  \bigstrut[t]\\
    LINE  & 0.345  & 0.123  & 1.334  & 1.946  & 0.213  & 0.403  & 0.411  & 0.281  & 2.115  & 3.411  & 1.912  & 7.502  & 0.323  & 0.261  & 2.470  & 4.632  & 3.015  & 15.102  \\
    Node2Vec & 0.397  & 0.177  & 1.282  & 1.889  & 0.218  & 0.403  & 0.430  & 0.315  & 2.349  & 4.185  & 2.846  & 8.031  & 0.376  & 0.285  & 2.149  & 4.137  & 2.877  & 15.089  \\
    DGI   & 0.395  & 0.151  & 1.279  & 1.848  & 0.214  & 0.398  & 0.449  & 0.313  & 1.866  & 2.989  & 2.731  & 7.990  & 0.377  & 0.295  & 2.218  & 4.149  & 2.767  & 15.148  \bigstrut[b]\\
    \hline
    RFN   & 0.383  & 0.117  & 1.234  & 1.773  & 0.215  & 0.413  & 0.432  & 0.283  & 1.967  & 3.213  & 1.914  & 7.576  & 0.389  & 0.304  & 2.235  & 4.453  & 3.234  & 15.085  \bigstrut[t]\\
    SRN2Vec & 0.384  & 0.124  & 1.126  & 1.707  & 0.198  & 0.395  & 0.443  & 0.307  & 1.803  & 3.111  & 1.989  & 7.459  & 0.396  & 0.318  & 2.265  & 4.307  & 3.002  & 15.098  \\
    HRNR  & 0.397  & 0.181  & 1.123  & 1.702  & 0.219  & 0.410  & 0.441  & 0.302  & 1.838  & 3.125  & 1.702  & 7.799  & 0.409  & 0.322  & 2.223  & 4.318  & 2.825  & 15.127  \\
    Toast & 0.462  & 0.201  & 1.166  & 1.674  & 0.201  & 0.394  & 0.457  & 0.319  & 1.822  & 2.958  & 1.692  & 7.505  & 0.439  & 0.378  & 2.193  & 4.224  & 2.735  & 15.086  \\
    JCLRNT & 0.497  & 0.218  & 1.149  & 1.753  & 0.208  & 0.395  & 0.493  & 0.331  & 1.819  & 3.083  & 1.594  & 7.506  & 0.443  & 0.401  & 2.182  & 4.123  & 2.686  & 15.104  \\
    HyperRoad & 0.499  & 0.226  & 1.101  & 1.672  & 0.104  & 0.337  & 0.504  & 0.339  & 1.665  & 2.952  & 1.585  & 7.502  & 0.473  & 0.442  & 2.298  & 4.305  & 2.696  & 15.081  \\
    SARN  & 0.479  & 0.221  & 1.062  & 1.666  & 0.188  & 0.397  & 0.506  & 0.344  & 1.582  & 2.949  & 1.583  & 7.444  & 0.476  & 0.456  & 2.092  & 4.115  & 2.525  & 15.131  \bigstrut[b]\\
    \hline
    \name & \textbf{0.530 } & \textbf{0.236 } & \textbf{1.015 } & \textbf{1.619 } & \textbf{0.095 } & \textbf{0.302 } & \textbf{0.525 } & \textbf{0.381 } & \textbf{1.531 } & \textbf{2.813 } & \textbf{1.536 } & \textbf{7.195 } & \textbf{0.496 } & \textbf{0.476 } & \textbf{1.985 } & \textbf{3.938 } & \textbf{2.365 } & \textbf{14.608 } \bigstrut\\
    \hline
    Improve & 6.21\% & 4.42\% & 4.43\% & 2.82\% & 8.65\% & 10.39\% & 3.75\% & 10.76\% & 3.22\% & 4.61\% & 2.97\% & 3.34\% & 4.20\% & 4.39\% & 5.11\% & 4.30\% & 6.34\% & 3.14\% \bigstrut\\
    \thickhline
    \end{tabular}%
    }
  \label{tab:road_result}%
  \vspace{-.2cm}
\end{table*}%

\begin{table*}[t]
  \centering
  \caption{Performance on Land Parcel-based Tasks.}
  \vspace{-0.2cm}
   \resizebox{\textwidth}{!}{
    \begin{tabular}{c|cc|cc|cc|cc|cc|cc|cc|cc|cc}
    \thickhline
    Data  & \multicolumn{6}{c|}{BJ}                       & \multicolumn{6}{c|}{XA}                       & \multicolumn{6}{c}{CD} \bigstrut\\
    \hline
    Tasks & \multicolumn{2}{c|}{Classification} & \multicolumn{2}{c|}{Flow prediction} & \multicolumn{2}{c|}{OD prediction} & \multicolumn{2}{c|}{Classification} & \multicolumn{2}{c|}{Flow prediction} & \multicolumn{2}{c|}{OD prediction} & \multicolumn{2}{c|}{Classification} & \multicolumn{2}{c|}{Flow prediction} & \multicolumn{2}{c}{OD prediction} \bigstrut\\
    \hline
    models\textbackslash{}metrics & Mi-F1 & Ma-F1 & MAE   & RMSE  & MAE   & RMSE  & Mi-F1 & Ma-F1 & MAE   & RMSE  & MAE   & RMSE  & Mi-F1 & Ma-F1 & MAE   & RMSE  & MAE   & RMSE \bigstrut\\
    \hline
    DeepWalk & 0.261  & 0.245  & 2.812  & 4.721  & 0.375  & 0.772  & 0.353  & 0.256  & 4.992  & 7.836  & 4.864  & 27.152  & 0.323  & 0.212  & 6.633  & 11.753  & 5.833  & 27.921 \bigstrut[t]\\
    LINE  & 0.231  & 0.219  & 3.051  & 4.923  & 0.391  & 0.797  & 0.329  & 0.183  & 5.016  & 9.062  & 5.517  & 27.459  & 0.215  & 0.203  & 7.626  & 12.825  & 6.236  & 27.941  \\
    Node2Vec & 0.287  & 0.256  & 2.723  & 4.623  & 0.373  & 0.753  & 0.391  & 0.279  & 4.902  & 7.441  & 4.843  & 26.203  & 0.344  & 0.273  & 6.167  & 10.032  & 6.105  & 28.087  \\
    DGI   & 0.292  & 0.283  & 2.539  & 4.478  & 0.363  & 0.744  & 0.378  & 0.263  & 4.711  & 7.612  & 4.455  & 20.111  & 0.341  & 0.281  & 6.202  & 11.541  & 5.912  & 28.554  \bigstrut[b]\\
    \hline
    HDGE  & 0.278  & 0.167  & 2.350  & 4.103  & 0.372  & 0.750  & 0.362  & 0.219  & 4.091  & 6.766  & 4.783  & 17.725  & 0.365  & 0.242  & 6.123  & 10.979  & 6.247  & 27.322  \bigstrut[t]\\
    ZE-Mob & 0.287  & 0.162  & 3.020  & 4.956  & 0.360  & 0.790  & 0.440  & 0.260  & 3.870  & 6.903  & 4.886  & 17.947  & 0.413  & 0.335  & 5.836  & 10.903  & 6.618  & 27.629  \\
    MV-PN & 0.296  & 0.243  & 2.791  & 4.652  & 0.358  & 0.771  & 0.387  & 0.279  & 3.907  & 6.702  & 4.912  & 18.021  & 0.334  & 0.289  & 6.124  & 10.758  & 5.916  & 27.932  \\
    MVURE & 0.327  & 0.221  & 2.344  & 4.402  & 0.367  & 0.723  & 0.422  & 0.283  & 3.591  & 6.481  & 4.517  & 17.921  & 0.425  & 0.326  & 5.331  & 10.240  & 5.860  & 27.523  \\
    Region2Vec & 0.311  & 0.243  & 2.291  & 4.123  & 0.391  & 0.738  & 0.431  & 0.302  & 3.623  & 6.479  & 4.551  & 17.329  & 0.423  & 0.339  & 5.213  & 10.198  & 5.931  & 27.623  \\
    MGFN  & 0.344  & 0.309  & 2.284  & 4.097  & 0.352  & 0.724  & 0.405  & 0.297  & 3.698  & 6.478  & 4.546  & 17.812  & 0.361  & 0.309  & 5.419  & 10.267  & 5.987  & 27.998  \\
    ReMVC & 0.339  & 0.313  & 2.279  & 4.089  & 0.383  & 0.722  & 0.442  & 0.311  & 3.584  & 6.483  & 4.512  & 17.033  & 0.432  & 0.342  & 4.976  & 9.772  & 5.811  & 27.396  \bigstrut\\
    \hline
    \name & \textbf{0.376 } & \textbf{0.343 } & \textbf{2.087 } & \textbf{3.764 } & \textbf{0.336 } & \textbf{0.686 } & \textbf{0.466 } & \textbf{0.331 } & \textbf{3.449 } & \textbf{6.238 } & \textbf{4.377 } & \textbf{16.408 } & \textbf{0.462 } & \textbf{0.377 } & \textbf{4.691 } & \textbf{9.402 } & \textbf{5.637 } & \textbf{26.461 } \bigstrut\\
    \hline
    Improve & 9.30\% & 9.58\% & 8.42\% & 7.95\% & 4.55\% & 4.99\% & 5.43\% & 6.43\% & 3.77\% & 3.70\% & 2.99\% & 3.67\% & 6.94\% & 10.23\% & 5.73\% & 3.79\% & 2.99\% & 3.41\% \bigstrut\\
    \thickhline
    \end{tabular}%
    }
  \label{tab:parcel_result}%
  \vspace{-.2cm}
\end{table*}%

\begin{table*}[t]
  \centering
  \caption{Performance on Trajectory-based Tasks.}
  \vspace{-0.2cm}
  \resizebox{\textwidth}{!}{
    \begin{tabular}{c|ccc|ccc|ccc|ccc|ccc|ccc}
    \hline
    Data  & \multicolumn{6}{c|}{BJ}                       & \multicolumn{6}{c|}{XA}                       & \multicolumn{6}{c}{CD} \bigstrut\\
    \hline
    Tasks & \multicolumn{3}{c|}{Travel time estimation} & \multicolumn{3}{c|}{Most similar trajectory search } & \multicolumn{3}{c|}{Travel time estimation} & \multicolumn{3}{c|}{Most similar trajectory search } & \multicolumn{3}{c|}{Travel time estimation} & \multicolumn{3}{c}{Most similar trajectory search } \bigstrut\\
    \hline
    models\textbackslash{}metrics & MAE   & RMSE  & MAPE  & MR    & HR@5  & HR@20 & MAE   & RMSE  & MAPE  & MR    & HR@5  & HR@20 & MAE   & RMSE  & MAPE  & MR    & HR@5  & HR@20 \bigstrut\\
    \hline
    trajectory2vec & 10.130  & 56.83  & 37.95  & 40.62  & 0.419  & 0.675  & 2.045  & 3.146  & 35.14  & 20.22  & 0.601  & 0.783  & 1.634  & 2.432  & 34.74  & 13.77  & 0.568  & 0.810  \bigstrut[t]\\
    t2vec & 10.030  & 56.65  & 36.42  & 31.01  & 0.483  & 0.741  & 2.028  & 3.126  & 33.73  & 19.56  & 0.621  & 0.893  & 1.629  & 2.426  & 34.45  & 12.82  & 0.636  & 0.865  \\
    Trember & 9.997  & 55.97  & 34.20  & 26.50  & 0.514  & 0.773  & 2.002  & 3.121  & 32.13  & 18.74  & 0.631  & 0.808  & 1.618  & 2.401  & 34.15  & 10.89  & 0.642  & 0.870  \\
    Toast & 10.681  & 47.73  & 52.63  & 27.75  & 0.505  & 0.705  & 2.154  & 3.264  & 33.93  & 21.37  & 0.646  & 0.802  & 1.710  & 2.491  & 37.24  & 13.20  & 0.645  & 0.868  \\
    JCLRNT & 10.218  & 46.49  & 49.18  & 23.15  & 0.536  & 0.806  & 2.166  & 3.246  & 33.12  & 11.18  & 0.651  & 0.870  & 1.690  & 2.481  & 36.42  & 12.76  & 0.653  & 0.870  \bigstrut[b]\\
    \hline
    DeepWalk & 11.737  & 57.67  & 56.72  & 41.59  & 0.469  & 0.703  & 2.338  & 3.365  & 37.53  & 19.80  & 0.562  & 0.811  & 1.862  & 2.554  & 35.12  & 21.45  & 0.592  & 0.853  \bigstrut[t]\\
    LINE  & 11.843  & 58.65  & 56.43  & 77.82  & 0.454 & 0.633  & 2.462  & 3.455  & 38.37  & 45.15  & 0.557  & 0.730  & 1.999  & 2.795  & 36.77  & 31.68  & 0.569  & 0.784  \\
    Node2Vec & 11.707  & 57.64  & 54.74  & 32.65  & 0.487  & 0.727  & 2.337  & 3.346  & 36.73  & 18.17  & 0.564  & 0.819  & 1.836  & 2.536  & 34.03  & 19.21  & 0.639  & 0.862  \\
    DGI   & 11.806  & 57.71  & 55.98  & 56.11  & 0.470 & 0.683  & 2.218  & 3.247  & 35.91  & 41.58  & 0.570  & 0.745  & 1.817  & 2.498  & 33.53  & 26.89  & 0.576  & 0.786  \bigstrut[b]\\
    \hline
    RFN   & 11.679  & 51.92  & 53.03  & 75.67  & 0.451  & 0.640  & 2.196  & 3.158  & 32.68  & 22.95  & 0.607  & 0.806  & 1.643  & 2.461  & 32.43  & 10.92  & 0.661  & 0.871  \bigstrut[t]\\
    SRN2Vec & 10.944  & 50.83  & 50.11  & 39.72  & 0.488 & 0.705  & 2.173  & 3.175  & 33.24  & 13.48  & 0.629  & 0.849  & 1.765  & 2.580  & 38.53  & 26.05  & 0.590  & 0.804  \\
    HRNR  & 10.418  & 51.10  & 52.13  & 43.13  & 0.491 & 0.703  & 2.154  & 3.185  & 33.51  & 16.53  & 0.606  & 0.825  & 1.671  & 2.413  & 34.90  & 17.45  & 0.610  & 0.851  \\
    HyperRoad & 10.172  & 39.11  & 39.37  & 35.39  & 0.445  & 0.697  & 1.926  & 3.056  & 31.63  & 12.36  & 0.624  & 0.860  & 1.604  & 2.394  & 33.13  & 9.79  & 0.654  & 0.865  \\
    SARN  & 9.446  & 36.97  & 32.59  & 18.04  & 0.557  & 0.826  & 2.087  & 3.060  & 34.65  & 10.77  & 0.664  & 0.879  & 1.594  & 2.405  & 31.89  & 9.54  & 0.663  & 0.877  \\
    \hline
    \name &  \textbf{8.940 } & \textbf{33.83 } & \textbf{30.37 } & \textbf{16.83 } & \textbf{0.610 } & \textbf{0.871 } & \textbf{1.863 } & \textbf{2.892 } & \textbf{29.32 } & \textbf{8.59 } & \textbf{0.707 } & \textbf{0.911 } & \textbf{1.469 } & \textbf{2.271 } & \textbf{30.35 } & \textbf{8.53 } & \textbf{0.691 } & \textbf{0.907 } \bigstrut\\
    \hline
    Improve & 5.36\% & 8.49\% & 6.81\% & 6.71\% & 9.52\% & 5.45\% & 3.27\% & 5.37\% & 7.30\% & 20.24\% & 6.48\% & 3.64\% & 7.84\% & 5.14\% & 4.83\% & 10.59\% & 4.22\% & 3.42\% \bigstrut\\
    \thickhline
    \end{tabular}%
    }
  \label{tab:traj_result}%
  \vspace{-.2cm}
\end{table*}%
\subsubsection{Task Settings}
To prevent data leakage, we exclude road or parcel labels when performing road or parcel label classification tasks. We employ 5-fold cross-validation to evaluate all methods' performance in road segment- and land parcel-based tasks. For tasks such as traffic flow prediction, OD flow prediction, and travel time prediction, we extract the ground truth values from the trajectories in the test set only. In the trajectory similarity search task, we draw 10,000 trajectories from the training set for fine-tuning the downstream model and 5,000 trajectories from the test set for testing the performance. Specifically, each trajectory needs to find the most similar trajectory among the other 4,999 ones, and we measure effectiveness using the Mean Rank (MR) and the top-k hit rate (HR).

\subsubsection{Evaluation Metrics}  
For the classification tasks, we utilize Micro-F1 (Mi-F1) and Macro-F1 (Ma-F1) as evaluation metrics for multi-classification tasks. For the flow prediction and OD flow prediction tasks (regression tasks), we employ mean absolute error (MAE) and root mean square error (RMSE) as evaluation metrics. For the travel time estimation task, we adopt three metrics: mean absolute error (MAE), mean absolute percentage error (MAPE), and root mean square error (RMSE). For the trajectory similarity search task, we evaluate the model using Mean Rank (MR), Hit Ratio at 5 (HR@5), and Hit Ratio at 20 (HR@20). 

\subsection{Performance Comparison}
The experimental results of the road segment-based tasks, land parcel-based tasks, and trajectory-based tasks are shown in Tab.~\ref{tab:road_result}, Tab.~\ref{tab:parcel_result}, Tab.~\ref{tab:traj_result} respectively. We repeat each experiment ten times and report the average results here. From the tables, we can make the following observations.

(1) General graph embedding methods, such as DeepWalk, LINE, Node2Vec, and DGI, exhibit poor performance in all tasks. Because they solely consider the graph topology and do not capture the complex semantics of city entities.

(2) For road segment-based tasks, early methods such as RFN, SRN2Vec, and HRNR primarily focus on considering simple attributes and geographic relationships of road segments, resulting in poor performance. Later work by Toast and JCLRNT attempt to incorporate trajectory features into the learning process, leading to improved performance. SARN emerges as the top-performing approach among these baseline models, and HyperRoad is the second best. SARN proposes graph contrastive learning to guide the model to learn the spatial structure of road networks. On the other hand, HyperRoad is designed to capture higher-order dependencies between roads by constructing hyperedges. However, they primarily focus on geographic adjacencies and ignore other connectivity relationships, such as function and mobility adjacencies. In contrast, our proposed \name model leverages multi-view graph construction, incorporating geographic, functional, and mobility relationships to comprehensively capture the interconnections between road segments and land parcels, leading to superior performance in various tasks.

(3) For land parcel-based tasks, HDGE and ZE-Mob incorporate geographic information of land parcels and dynamics of trajectories based on Node2Vec, leading to significantly improved performance. On the other hand, multi-view representation learning methods like MVURE, MGFN, and ReMVC achieve better results as they capture richer information from multiple views constructed on the data. Among them, ReMVC stands out as it utilizes contrastive learning based on the multi-view representation learning approach, achieving the best results in experiments. However, ReMVC does not consider the graph structure of land parcels and only constructs the representation of the land parcels using POI attributes and human mobility attributes for contrastive learning. In contrast, our proposed \name model not only captures the graph structure information of urban areas but also utilizes multi-view representation learning and contrastive learning to achieve superior performance in both road segment-based and land parcel-based tasks.

(4) For the trajectory-based tasks, the first class of models trained on learned road segment representations shows similar results to the road segment-based tasks, with SARN continuing to be the top-performing baseline model. The models trained based on trajectory representations, such as t2vec, trajectory2vec, and Trembr, outperform most models trained based on road segment representations. This performance improvement can be attributed to the fact that the trajectory representation models allow for fine-tuning of both upstream and downstream components during testing. In contrast, the models with fixed road segment representations only take static representation vectors as inputs.

(5) Despite the good performance of the baselines, our proposed \name outperforms all of them across all tasks and metrics on the three datasets, as confirmed by the Student's t-test at level 0.01. This demonstrates the effectiveness of our approach in multi-type map entity representation learning. The superior performance can be attributed to the ability of our model to integrate road segment and land parcel representations in a unified manner. Moreover, introducing two-level contrastive learning tasks in our model allows us to capture fine-grained and high-level relationships within the \home graph. 


\begin{figure}[t]
\centering
\includegraphics[width=0.95\columnwidth]{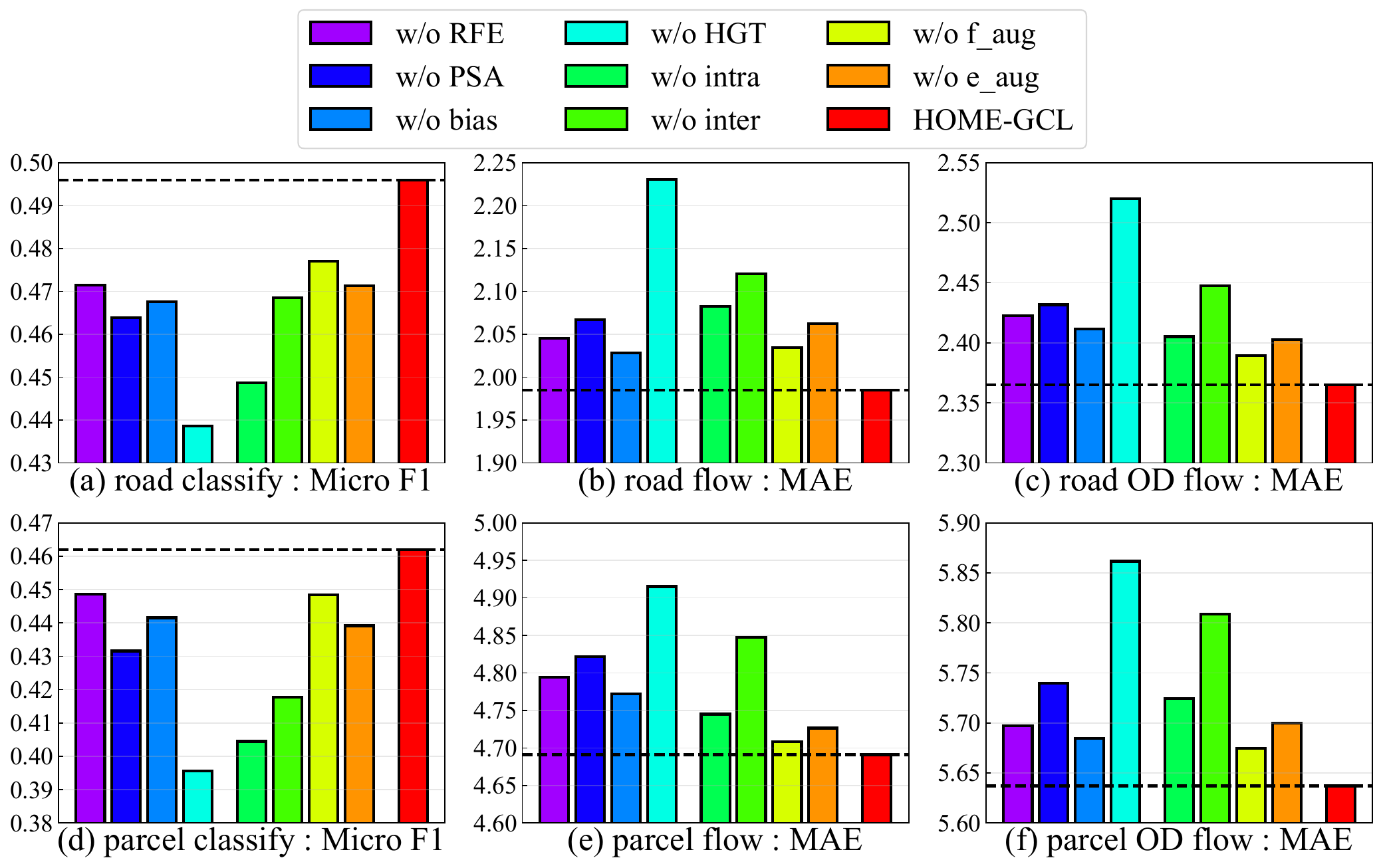}
    \vspace{-.3cm}    
    \caption{{Ablation Study on the \cd Dataset.}}
    \label{fig:abla}
    \vspace{-.4cm}
\label{exp:abla}
\end{figure}

\subsection{Ablation Study}\label{exp:ablation}
In order to investigate the contributions of each sub-module in \name, we conduct ablation experiments on all datasets. We repeat each experiment ten times and report the average results in Fig.~\ref{fig:abla}. Due to limited space, we only present the results for the six subtasks based on road segments on the \cd dataset and similar results are observed for other datasets.



\subsubsection{Impact of basic sub-modules} (a) \textit{w/o \gf}, which removes the Raw Entity Feature Encoding module, (b) \textit{w/o \psa}, which removes the Parcel-segment Shape Attention module, (c) \textit{w/o bias}, which uses Eq.~\eqref{eq:att} instead of Eq.~\eqref{eq:att_enchance} to calculate the attention score for the \psa module, and (d) \textit{w/o \hgt}, which removes the Heterogeneous Graph Transformer module. 

The findings suggest that removing the \gf module significantly hampers representation learning, highlighting the importance of the underlying features. The omission of the \psa module has a more pronounced impact on both task levels, particularly for land parcels, underscoring the significance of geometric structure features and cross-entity learning. When the bias term in the attention equation is eliminated, the model disregards the shape feature of parcel polygons and their relative positional relationship with surrounding road segments, resulting in a detrimental effect on performance. Furthermore, excluding the \hgt module, which addresses heterogeneity in the \home, yields the poorest performance as it solely relies on the entity representation output from \psa. In fact, the \psa module can be viewed as an extension of the \hgt module, incorporating additional attentional computations to explicitly model the spatial geometric distribution and shape relationships between road segments and land parcels, which are not accounted for in the standard \hgt model.



\subsubsection{Impact of contrastive learning} (a) \textit{w/o intra}, which uses only inter-entity contrast, (b) \textit{w/o inter}, which uses only intra-entity contrast, (c) \textit{w/o f\_aug}, which removes the node feature augmentation, and (d) \textit{w/o e\_aug}, which removes the edge augmentation. 

Removing intra-entity contrast impacts more on classification tasks while removing inter-entity contrast impacts more on regression tasks. This is because intra-entity contrast learning helps distinguish representations of entities of the same category, which is more beneficial for the classification task of urban entities. On the other hand, inter-entity contrast learning helps co-learn entities of different categories, which is more beneficial for the flow prediction task. The interactions between land parcels and road segments affect each other, illustrating the importance of co-learning both entities together. Finally, removing edge-based augmentation is more effective than removing feature-based augmentation as we consider multi-view edges in the \home.


\begin{figure}[t]
\centering
\includegraphics[width=0.95\columnwidth]{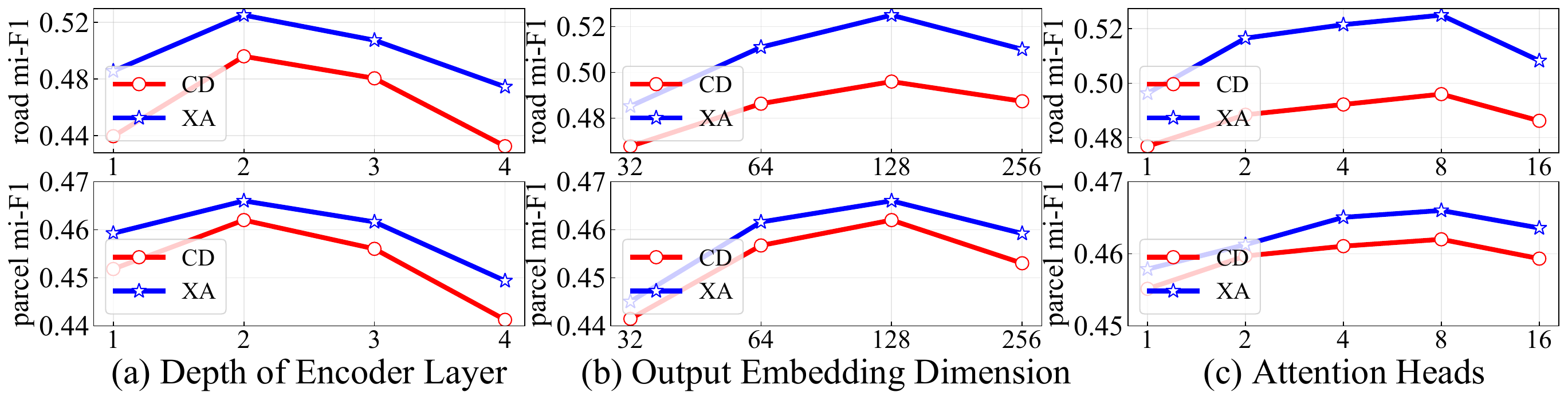}
    \vspace{-.3cm}
    \caption{Parameter Sensitivity Analysis.}
    \label{fig:para}
    \vspace{-.3cm}
\end{figure}

\subsection{Parameter Sensitivity}\label{exp:param}
We conduct a parameter sensitivity analysis for critical hyperparameters, such as encoder layers $L$, embedding size $D$, and attention heads $H$ on the \cd and \xa datasets. We present the results of two classification tasks on these datasets, as the results of other datasets and tasks are similar. From Fig.~\ref{fig:para}, we observe that the model performance initially improves with an increase in $L$ and $D$, but when they become too large, the performance starts to deteriorate due to overfitting. As for the attention heads, the model's performance increases with an increase in $H$, indicating that more diverse information can be introduced to stabilize the training process, leading to improved performance. However, too many attention heads can also negatively impact the model's performance.

\begin{figure}[t]
\centering
\includegraphics[width=0.95\columnwidth]{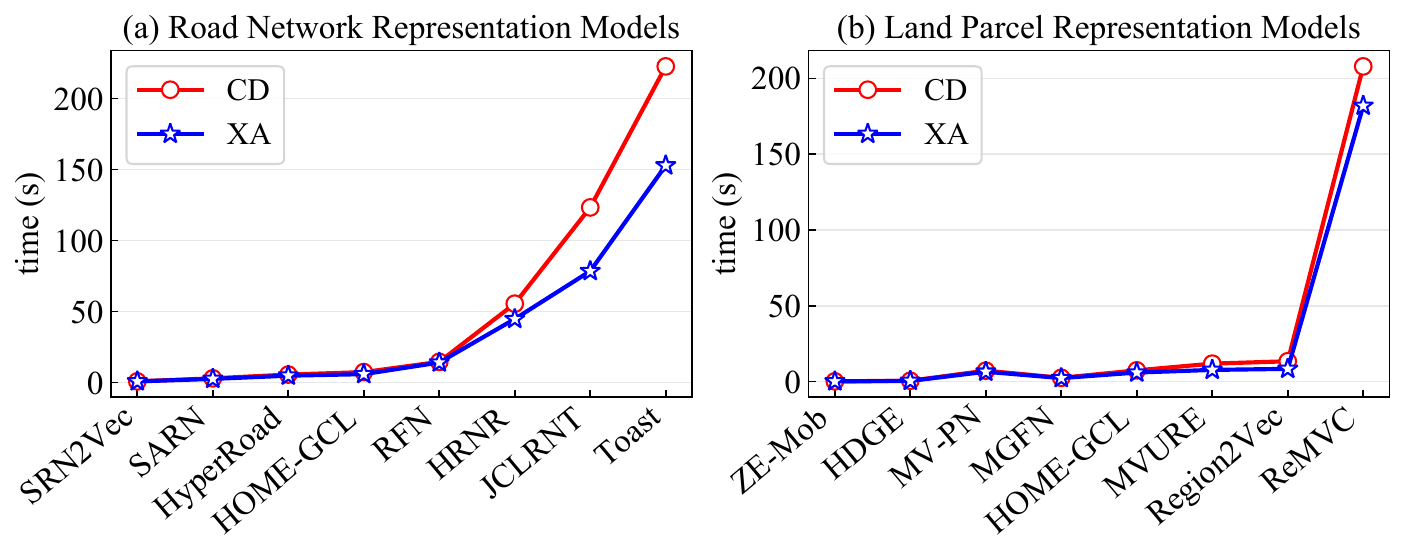}
    \vspace{-.4cm}
    \caption{Model Training Time on \xa and \cd Datasets.}
    \label{fig:effe}
    \vspace{-.5cm}
\end{figure}

\subsection{Model Efficiency And Scalability}\label{exp:efficiency}
Fig.~\ref{fig:effe} shows the training time results for road segment and land parcel representation models. We train these models separately on an NVIDIA GeForce 3090 GPU and report the average time for one training epoch. We can see that some simpler models, such as HDGE, ZE-Mob, and SRN2Vec, have shorter training times, but their performance is also inferior. Among the road network representation models, JCLRNT and Toast have longer training times because they use Transformers to model trajectories and obtain road representations, which is time-consuming. For the land parcel representation models, ReMVC has a long training time because it processes only one land parcel per iteration during training, resulting in linear growth in training time with the number of parcels.

Although our model has a slightly longer training time than some of the better-performing baselines, such as HyperRoad, SARN, and MGFN, our model acquires both road segment and land parcel-level representations synchronously. In contrast, the baselines only acquire single-scale representations. Joint learning enables mutual reinforcement of representation learning, achieving trade-offs between performance and efficiency.



\subsection{Case Study}\label{exp:vis}

\begin{figure}[t]
    \centering
    \subfigure[\name (parcel)]{
        \includegraphics[width=0.28\columnwidth]{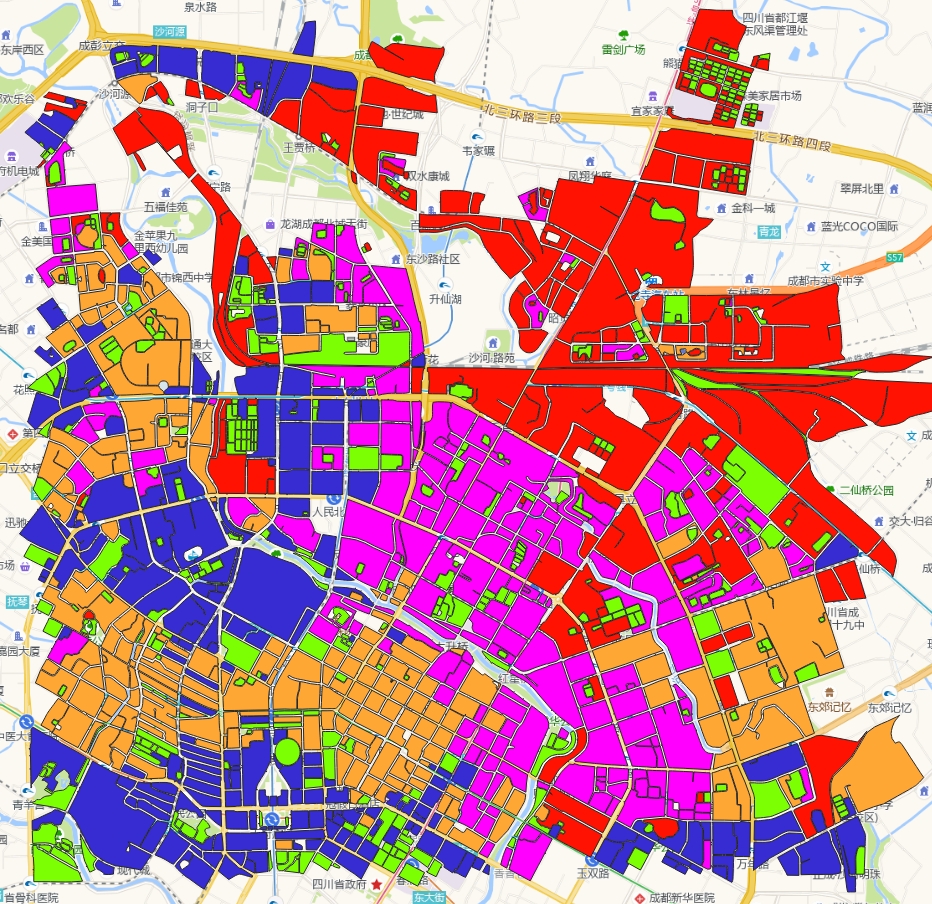}
         \label{fig:cluster_a}
    }
    \subfigure[DeepWalk (parcel)]{
        \includegraphics[width=0.28\columnwidth]{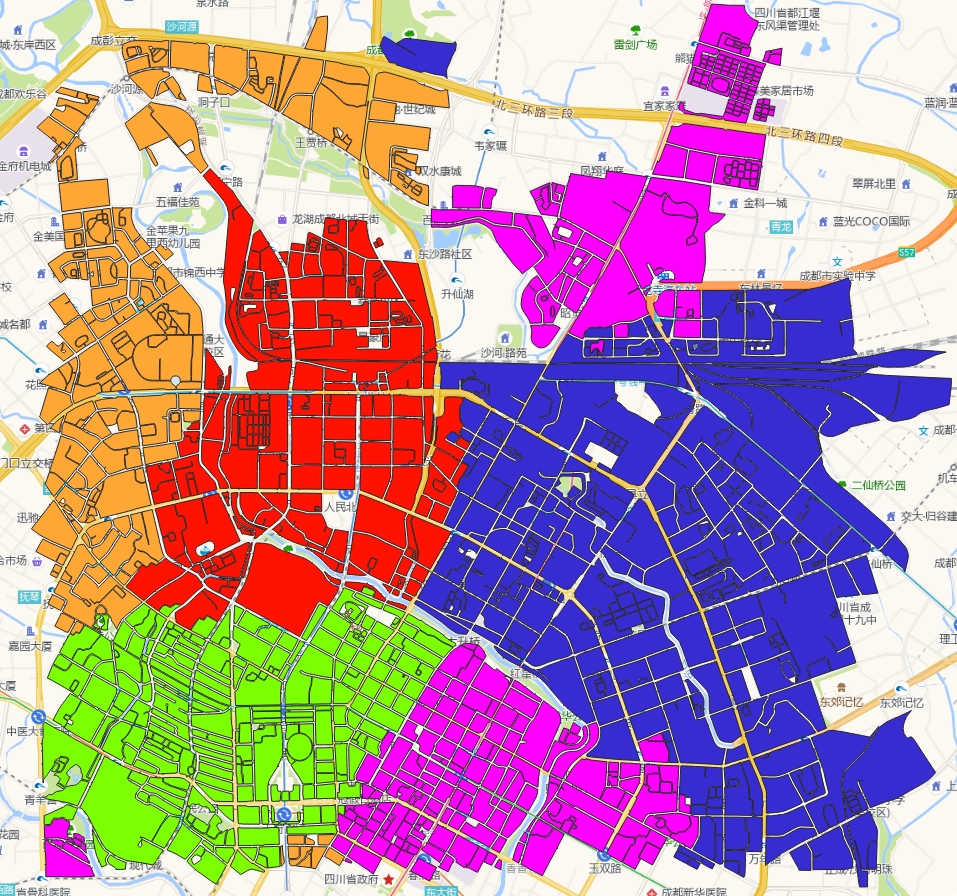}
         \label{fig:cluster_b}
    }
    \subfigure[ReMVC (parcel)]{
        \includegraphics[width=0.28\columnwidth]{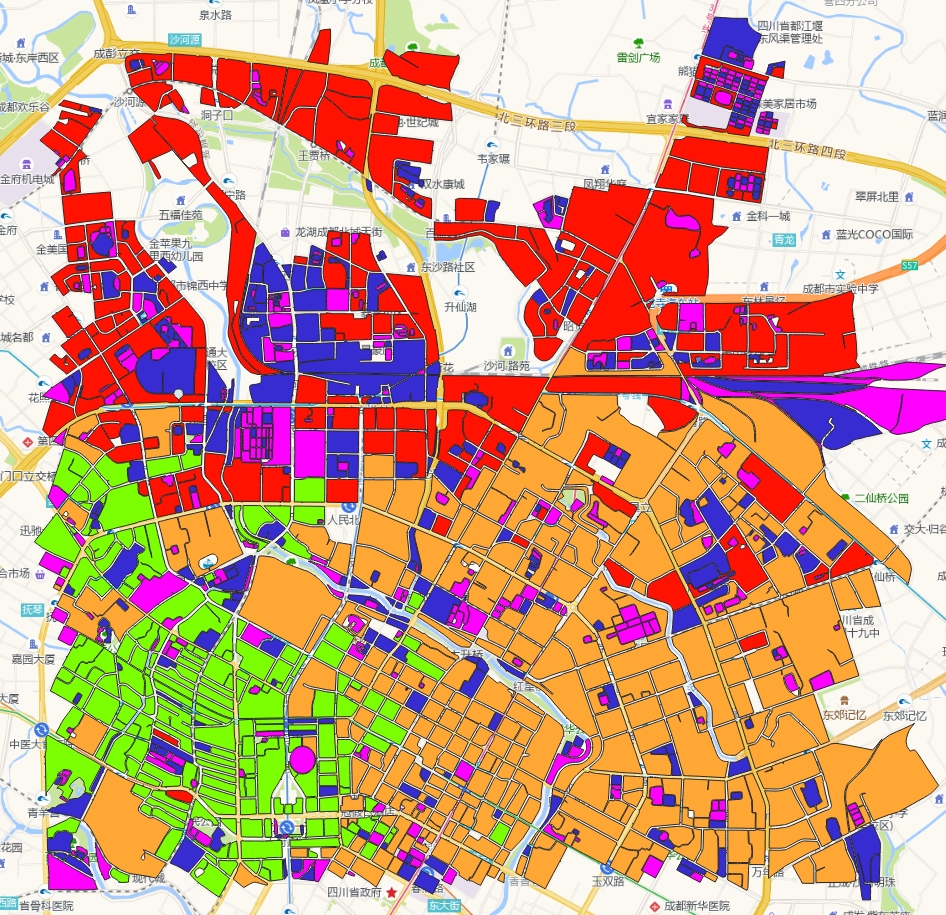}
         \label{fig:cluster_c}
    }
    \\
    \subfigure[\name (road)]{
        \includegraphics[width=0.28\columnwidth]{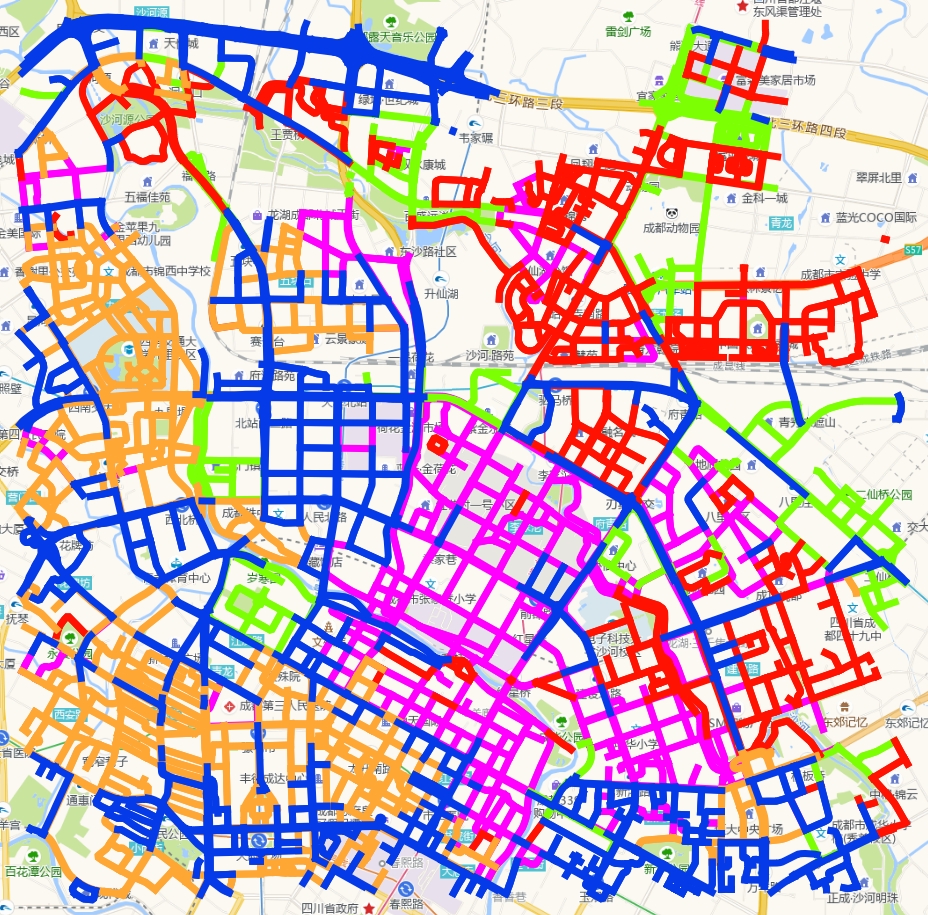}
         \label{fig:cluster_d}
    }
    \subfigure[DeepWalk (road)]{
        \includegraphics[width=0.28\columnwidth]{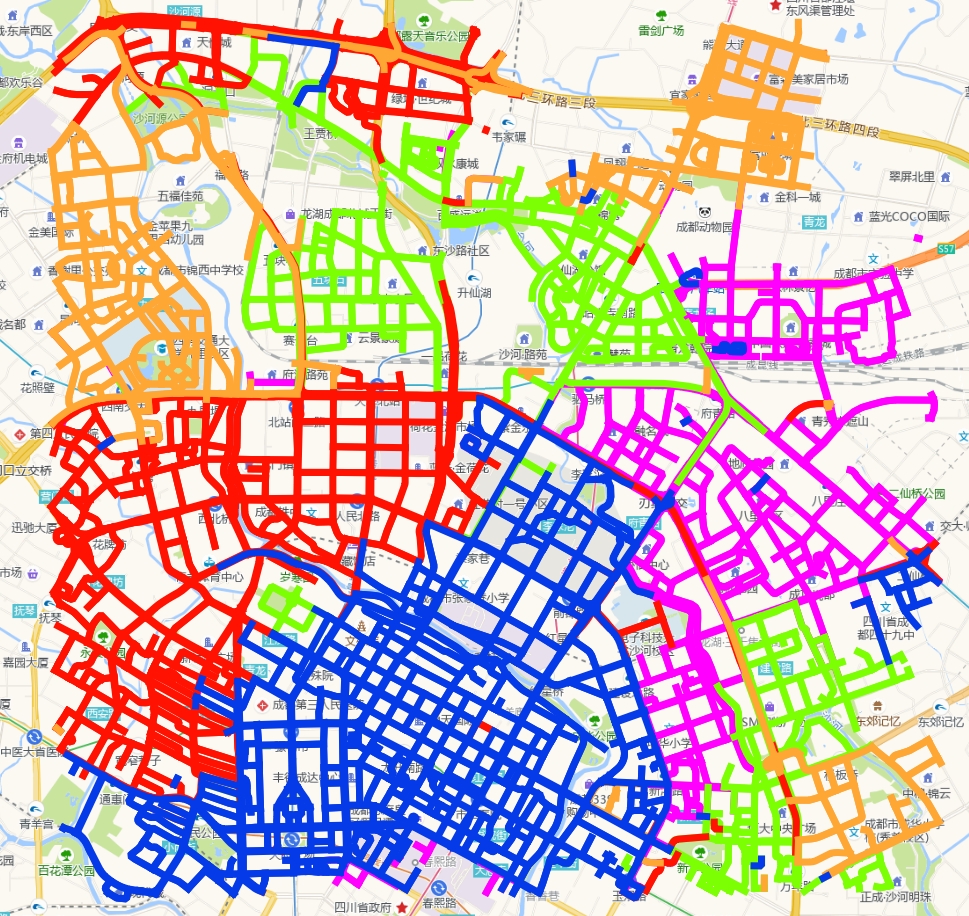}
         \label{fig:cluster_e}
    }
    \subfigure[SARN (road)]{
        \includegraphics[width=0.28\columnwidth]{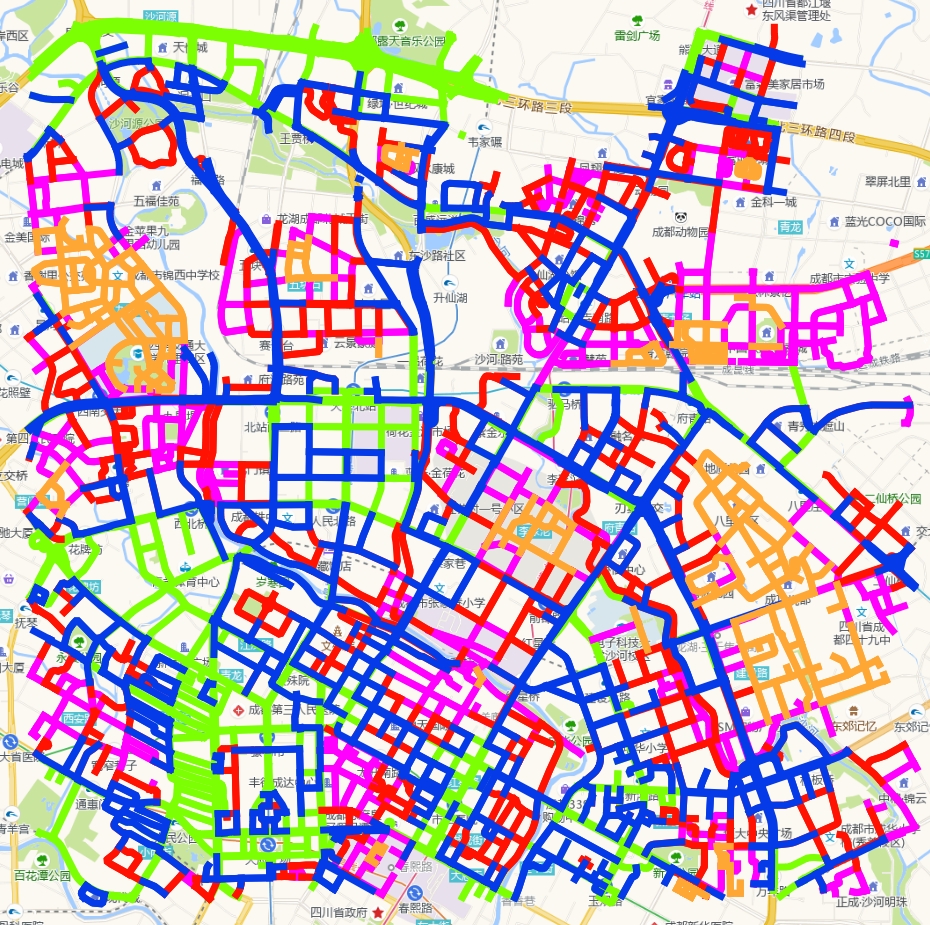}
         \label{fig:cluster_f}
    }
    \\
    \subfigure[\name (ideal)]{
        \includegraphics[width=0.28\columnwidth]{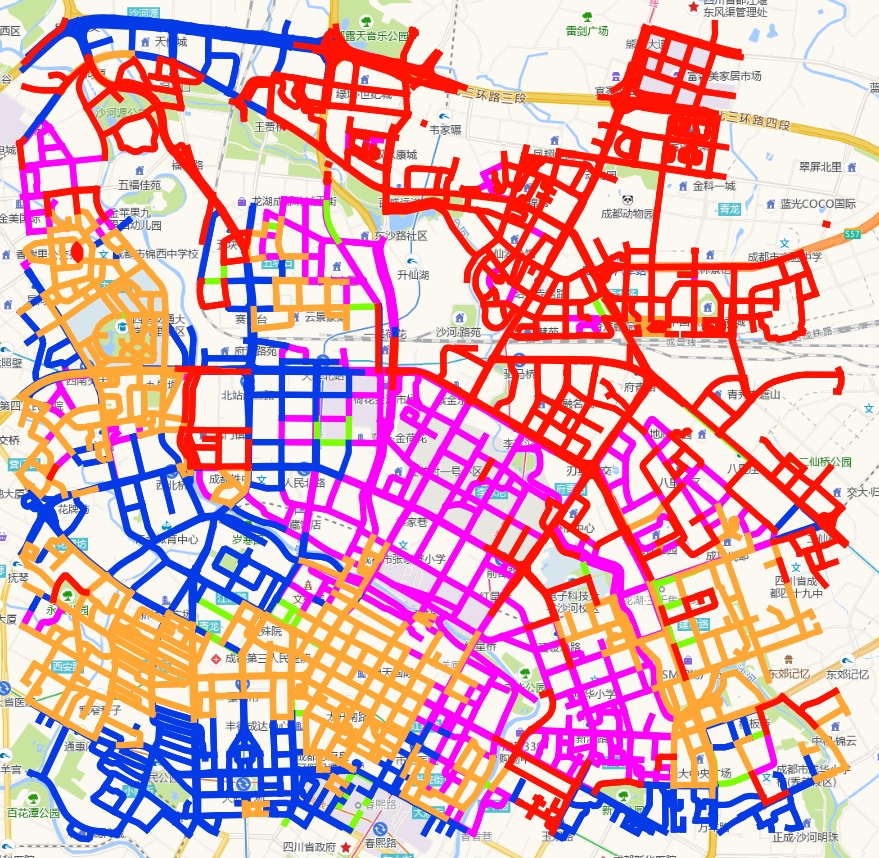}
         \label{fig:cluster_g}
    }
    \subfigure[DeepWalk (ideal)]{
        \includegraphics[width=0.28\columnwidth]{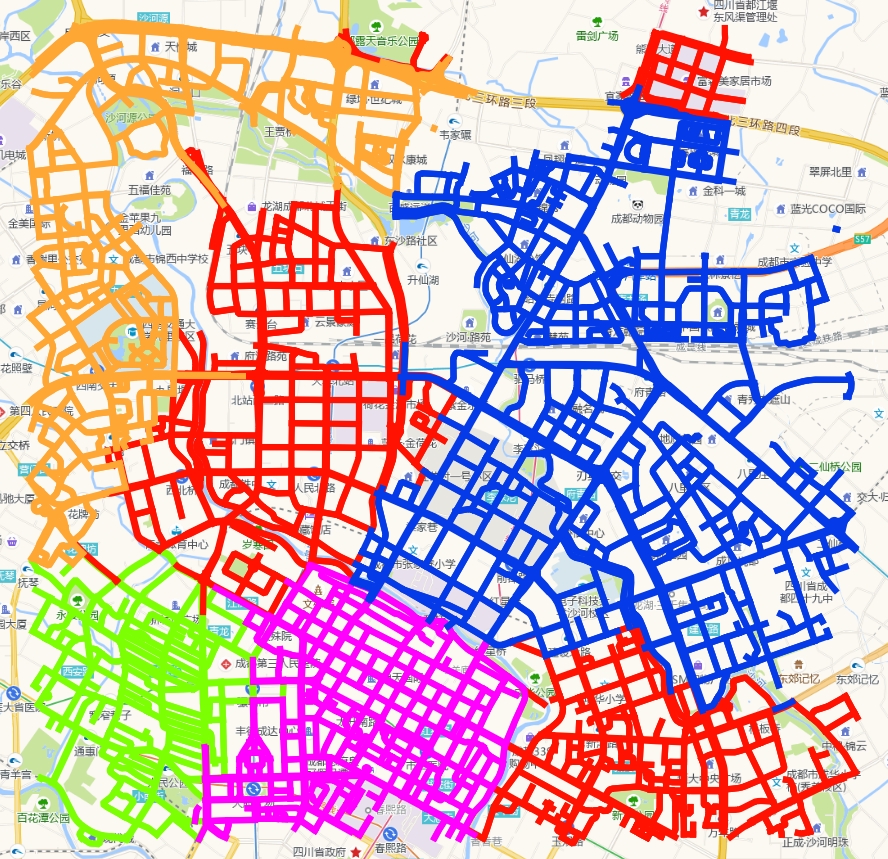}
         \label{fig:cluster_h}
    }
    \subfigure[SARN (ideal)]{
        \includegraphics[width=0.28\columnwidth]{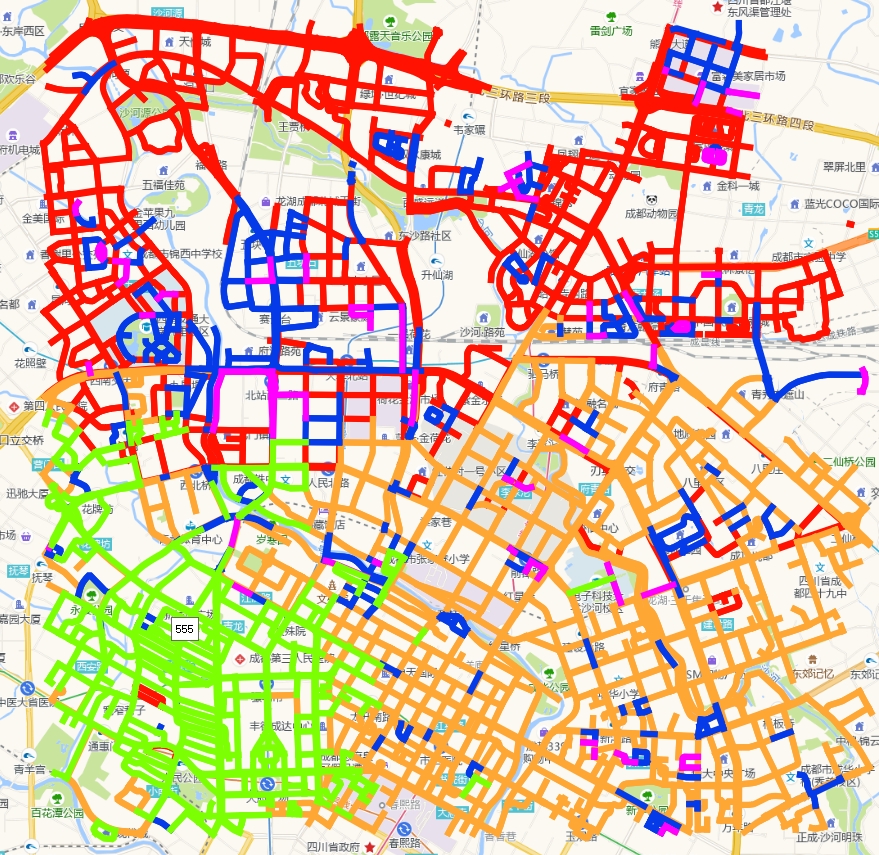}
         \label{fig:cluster_i}
    }
    \vspace{-.1cm}
    \caption{{Clustering of Learned Road Segment and Land Parcel Representations.}}
    \label{fig:cluster}
    \vspace{-.3cm}
\end{figure}

The \name model is designed for joint learning of road segment and parcel representations using a unified model. An advantage of this method is that the learned representations for heterogeneous entities are under the same system. For a segment and a parcel that are adjacent or have similar functions, their representations generated by our model should have a certain consistency, while those generated by a separate model cannot ensure this feature. To verify this point, we demonstrate the clustering experiments on learned representations in Fig.~\ref{fig:cluster}. In this experiment, we take the Chengdu city dataset as an example and perform $k$-means clustering with $k$=5 on the parcel and road segment representations, respectively. We plot the parcels and segments within the same cluster as the same color. 

As a comparison, we adopt two groups of baselines. In the first group of baseline, we use the DeepWalk model, which is a general graph node representation model, to generate the representations of parcels and segments, respectively, \ie ``one model for each of the two categories''. In the second group of baselines, we use ReMVC to generate representations of parcels and SARN to generate representations of segments. The two models are specifically designed for corresponding map entities, \ie ``two models for each of the two categories''. Correspondingly, our model is ``one model for both of the two categories''.

The clustering results of the parcel representations learned by the model are depicted in plots (a), (b), and (c), while plots (d), (e), and (f) display the clustering results of the road segment representations learned by the model. Additionally, plots (g), (h), and (i) illustrate the clustering results obtained by utilizing the assignment matrix $\bm{A}^{SR} \in \mathbb{R}^{N_S \times N_R}$ for roads and parcels. We map the clustering of parcels to the corresponding clustering of road segments associated with those parcels. In other words, plots (g), (h), and (i) represent the ideal case where road segment clustering results are obtained.

As can be seen in the figures, the clustering results of our model show a very high consistency in the geographical distribution of the clusters for the different categories of entities \ie the road segments corresponding to land parcels grouped together spatially are also clustered together (see Fig.~\ref{fig:cluster_d} and Fig.~\ref{fig:cluster_g}). In contrast, DeepWalk generates separate representations for road segments and land parcels (Fig.\ref{fig:cluster_e} and Fig.\ref{fig:cluster_h}), resulting in clustering without consistency between the two entities. The representations learned by the optimal baselines, \ie ReMVC and SARN also lack consistency (see Fig.~\ref{fig:cluster_f} and Fig.~\ref{fig:cluster_i}). 

We further quantitatively measure this consistency using two road segment clustering methods: plots (d), (e), and (f) represent one method, while plots (g), (h), and (i) represent another method. To assess the degree of similarity or consistency between these two clusters, we compute the Normalized Mutual Information (NMI) and Adjusted Rand Score (ARS). The experimental results are presented in Tab.~\ref{tab:cluster_quantitative}, where higher values indicate a higher degree of similarity and consistency between segment and parcel representations. It is evident that our \name significantly outperforms the two baselines in terms of NMI and ARS scores.

\begin{table}[t]
  \centering
  \caption{Results of Representation Consistency Experiments.}
  \resizebox{\columnwidth}{!}{
    \begin{tabular}{cccc|c|c|c|c|c|c}
    \thickhline
    \multicolumn{4}{c|}{Data}     & \multicolumn{2}{c|}{BJ} & \multicolumn{2}{c|}{XA} & \multicolumn{2}{c}{CD} \bigstrut\\
    \hline
    \multicolumn{4}{c|}{Models\textbackslash{}Metrics} & NMI   & ARS   & NMI   & ARS   & NMI   & ARS \bigstrut[t]\\
    \hline
    \multicolumn{4}{c|}{\name+ \name} & \textbf{0.194}  & \textbf{0.143}  & \textbf{0.307}  & \textbf{0.254}  & \textbf{0.400}  & \textbf{0.336}  \\
    \multicolumn{4}{c|}{DeepWalk + DeepWalk} & 0.009  & 0.005  & 0.129  & 0.059  & 0.203  & 0.134  \\
    \multicolumn{4}{c|}{ReMVC + SARN} & 0.015  & 0.008  & 0.114  & 0.073  & 0.037  & 0.024  \bigstrut[b]\\
    \thickhline
    \end{tabular}%
    }
  \label{tab:cluster_quantitative}%
  \vspace{-0.5cm}
\end{table}%

\section{Related Work}

\subsection{Map Entity Representation Learning}

Map entity representation learning (MERL) has emerged as a powerful approach for modeling urban city map data and generating generic representation vectors. These vectors hold significant value for a wide range of complex tools and tasks, particularly in the current era of pre-trained foundational models, where the significance of urban map entity data representation learning is amplified, as highlighted by recent research CityFM~\cite{balsebre2023cityfm}. The urban map entities encompass points of interest (POIs), road segments, and land parcels, each of which is discussed independently in this section. 

\subsubsection{Point of Interest Representation Learning}
Point of Interest (POI) representation learning typically involves capturing users' mobility patterns and generating POI representations based on their check-in sequence data from location-based social networks. Earlier approaches utilized generic word representation learning methods~\cite{word2vec} to learn POI representations from check-in sequence data~\cite{poi2vec, poiskipgram}. More recent work, such as HMRM~\cite{chen2020modeling}, incorporates POI co-occurrences and POI-time co-occurring in sequences along with implicit matrix factorization to learn representations. Deep learning methods have also played a crucial role in modeling check-in data. Researchers often employ sequence models like LSTMs~\cite{LSTM} to capture temporal patterns from POI check-in sequences and generate representations for predicting the next POI~\cite{yu2020category, kong2018hst}. 

\subsubsection{Road Segment Representation Learning}
Road segment representation learning has undergone significant developments in three stages. Initially, researchers relied on topology-aware graph embedding methods like DeepWalk~\cite{deepwalk} and Node2Vec~\cite{Node2Vec}, but these methods had limitations in capturing complex road properties and traffic semantics. Subsequently, more graph-based road network representation methods have emerged with the progress of graph neural networks and representation learning techniques. These approaches consider topology and geospatial attribute information~\cite{RFN, IRN2Vec, SRN2Vec, HRNR, Toast, HyperRoad}. In particular, HRNR~\cite{HRNR} clusters road segments as structure regions and function zones and learns representations of individual segments and segment communities simultaneously. However, HRNR is still a model for homogenesis map entity representation learning. In the latest developments, two self-supervised methods, JCLRNT~\cite{JCLRNT} and SARN~\cite{SARN}, have been proposed. However, they primarily focus on road network features and overlook the interplay between urban entities at different scales. 

\subsubsection{Land Parcel (Region) Representation Learning}
Early methods for land parcel representation learning utilized skip-gram methods~\cite{word2vec} to learn embeddings of land parcels from sequences of human mobility data~\cite{HDGE, ZE-Mob}. Subsequent studies focused on modeling land parcel features using multiple views of data, such as geographic distances and human mobility data~\cite{CGAL, MV-PN}. More recently, with the advancement of graph neural networks, researchers have explored the graph structure of multiple views of land parcels for representation learning~\cite{MVURE, GEML, Region2Vec, MGFN}. In the most recent research, the ReMVC model~\cite{ReMVC} proposes a self-supervised representation learning scheme incorporating contrastive learning across views, demonstrating promising results. However, it does not consider the graph structure of land parcels in the city.


\subsection{Graph Contrastive Learning}
Contrastive learning is a self-supervised learning technique that learns representations by contrasting positive and negative samples~\cite{simclr, moco}. It maximizes similarity between representations of positive samples and minimizes similarity with negative samples. Graph contrastive learning is an application of this technique to graph-structured data. Graph contrastive learning has seen increasing interest, and several related works have explored its potential in learning meaningful representations from graph-structured data. For instance, Deep Graph Infomax (DGI)\cite{DGI} contrasts nodes and high-level graph summaries to learn graph representations. Similar to research in other fields~\cite{simclr, Consert}, data augmentation plays a crucial role in graph contrastive learning. MVGRL~\cite{MVGRL} leverages node diffusion and contrastive learning for node and graph-level representations. GraphCL~\cite{GCL} designs four types of graph augmentations to incorporate various priors. GCA~\cite{GCA} introduces adaptive data augmentation on both topology and attribute levels of the graph. These studies collectively highlight the growing interest in graph contrastive learning and its potential for learning meaningful representations from graph-structured data. 


\section{Conclusion and Future work}
This paper introduces \name, a novel framework that unifies the representation learning of road networks and land parcels. The key component of \name is the heterogeneous map entity graph (\home graph), which organizes the two categories of map entities, namely road segments and land parcels. To generate representation vectors, we propose a \home encoder that incorporates parcel-segment joint feature encoding and a heterogeneous graph transformer. Furthermore, we introduce two contrastive learning tasks to capture the interplay between different map entities. Extensive experiments on three large-scale datasets demonstrate the effectiveness of \name in various map entity representation learning tasks. The proposed approach has significant implications for advancing our understanding of urban dynamics and supporting applications in urban planning and transportation. 

In future work, we plan to enhance \name by incorporating POIs into the \home, enabling joint training and representation of POI-segment-parcel interactions, and empowering a wide range of urban data analysis and management tasks. We also aim to explore the potential of incorporating additional data sources, such as satellite imagery and social media, to further enrich the representation of urban entities.

\ifCLASSOPTIONcaptionsoff
  \newpage
\fi



%

\bibliographystyle{IEEEtran}
\bibliography{sample}




%

\begin{IEEEbiography}{Jiawei Jiang}
Jiawei Jiang is a current master's student at the School of Computer Science, Beihang University (BUAA). His main research interests are spatial-temporal data mining and spatial-temporal representation learning, and he has published relevant papers in AAAI, ICDE, ACM SIGSpatial and other conferences.
\end{IEEEbiography}

\begin{IEEEbiography}{Yifan Yang}
Yifan Yang is a current master's student at the School of Computer Science, Beihang University (BUAA), and her main research interest is spatial-temporal representation learning.
\end{IEEEbiography}

\begin{IEEEbiography}{Jingyuan Wang}
Jingyuan Wang received the PhD degree from the Department of Computer Science and Technology, Tsinghua University, Beijing, China. He is currently an associate professor of the School of Computer Science and Engineering, Beihang University, Beijing, China. He is also the head of the Beihang Interest Group on SmartCity (BIGSCity), and vice director of the Beijing City Lab (BCL). He published more than 30 papers on top journals and conferences, as well as named inventor on several granted US patents. His general area of research is data mining and machine learning, with special interests in smart cities.
\end{IEEEbiography}

\begin{IEEEbiography}{Junjie Wu}
Junjie Wu received the Ph.D. degree in management science and engineering from the School of Economics and Management, Tsinghua University, Beijing, China, in 2008. He is currently a Full Professor with the Information Systems Department, School of Economics and Management, Beihang University, Beijing, China. He is also the Director of the Research Center for Data Intelligence (DIG), the Vice Director of Beijing Key Laboratory of Emergency Support Simulation Technologies for City Operations, and a Senior Researcher with the Beijing Innovation Center for Big Data and Brain Computing. His research interests include is data mining and machine learning, with a special interest in social, urban and financial computing. Prof. Wu is the recipient of various nation-wide academic awards in China, including the NSFC Distinguished Young Scholars, the MOE Changjiang Young Scholars, and the National Excellent Doctoral Dissertation.
\end{IEEEbiography}






\end{document}